\documentclass[journal]{IEEEtran}
\usepackage{algorithm}
\usepackage{cuted}
\usepackage{algorithmic}
\usepackage{mathtools}
\usepackage{caption}
\usepackage{cite}
\usepackage{amsthm}
\usepackage{graphicx}
\usepackage{textcomp}
\usepackage[english]{babel}
\usepackage{amsfonts}
\usepackage{amsbsy}
\usepackage{indentfirst}
\usepackage{subfigure}
\usepackage{epstopdf}
\usepackage{amsmath}
\usepackage{times}
\usepackage{latexsym}
\usepackage{bm}
\usepackage{amssymb}
\usepackage{stfloats}
\usepackage{cases}
\usepackage{array}
\usepackage{setspace}
\usepackage{fancyhdr}
\usepackage{amscd}
\usepackage{enumerate}
\usepackage{multirow}
\usepackage{float}
\usepackage{xcolor}
\usepackage{longtable}
\usepackage{clrscode}

\usepackage{enumitem}
\usepackage{listings}
\usepackage{verbatim}
\usepackage{blindtext}
\newtheorem{Theorem}{Theorem}
\newtheorem{Lemma}{Lemma}
\newcommand{\defeq}{\vcentcolon=}
\newcommand{\indep}{\perp \!\!\! \perp}
\newcommand\itema{\item[\textbf{A1:}]}
\newcommand\itemb{\item[\textbf{A2:}]}
\newcommand\itemc{\item[\textbf{A3:}]}
\newcommand\itemSa{\item[\textbf{S1:}]}
\newcommand\itemSb{\item[\textbf{S2:}]}
\newcommand\itemSc{\item[\textbf{S3:}]}
\newcommand\itemSd{\item[\textbf{S4:}]}

\begin{document}

\title{Gaussian Process Upper Confidence Bounds in Distributed Point Target Tracking over Wireless Sensor Networks}


\author{\IEEEauthorblockN{Xingchi Liu\IEEEauthorrefmark{1},
Lyudmila Mihaylova\IEEEauthorrefmark{1},
Jemin George\IEEEauthorrefmark{2},
Tien Pham\IEEEauthorrefmark{3}
}\\ 
\IEEEauthorblockA{
\IEEEauthorrefmark{1}University of Sheffield, UK\\ 
\IEEEauthorrefmark{2}DEVCOM Army Research Laboratory, USA\\
\IEEEauthorrefmark{3}MITRE Labs, USA\\
Email: {xingchi.liu@sheffield.ac.uk}
{l.s.mihaylova@sheffield.ac.uk}
{jemin.george.civ@army.mil},
{tienpham@mitre.org}}
\thanks{We are grateful to the support for the UK Dstl and US MoD SIGNeTS project. Research was sponsored by the US Army Research Laboratory and the UK MOD University Defence Research Collaboration (UDRC) in Signal Processing, and was accomplished under Cooperative Agreement Number W911NF-20-2-0225. The views and conclusions contained in this document are those of the authors and should not be interpreted as representing the official policies, either expressed or implied, of the Army Research Laboratory, the MOD, the U.S. Government or the U.K. Government. The U.S. Government and U.K. Government are authorized to reproduce and distribute reprints for Government purposes notwithstanding any copyright notation herein. We acknowledge the support from the EPSRC through EP/T013265/1 project NSF-EPSRC: ``ShiRAS. Towards Safe and Reliable Autonomy in Sensor Driven Systems'' and the support for ShiRAS by the USA National Science Foundation under Grant NSF ECCS 1903466. For the purpose of open access, the author has applied a Creative Commons Attribution (CC BY) licence to any Author Accepted Manuscript version arising. }
}

\markboth{IEEE Journal of Selected Topics in Signal Processing}{ \MakeLowercase{\textit{Liu and Mihaylova}}: Gaussian Process Upper Confidence Bounds in Distributed Point Target Tracking over Wireless Sensor Networks}


\maketitle

\begin{abstract}
\color{black}Uncertainty quantification plays a key role in the development of autonomous systems, decision-making, and tracking over wireless sensor networks (WSNs). However, there is a need of providing uncertainty confidence bounds, especially for distributed machine learning-based tracking, dealing with different volumes of data collected by sensors. This paper aims to fill in this gap and proposes a distributed Gaussian process (DGP) approach for point target tracking and derives upper confidence bounds (UCBs) of the state estimates. A unique aspect of the proposed approach consists in the theoretical results are derived that prove its maximum accuracy for tracking with and without clutter measurements. Particularly, the developed approaches with uncertainty bounds are generic and can provide trustworthy solutions with an increased level of reliability. A novel hybrid Bayesian filtering method is proposed to enhance the DGP approach by adopting a Poisson measurement likelihood model. The proposed approaches are validated over a WSN case study, whose sensors have limited sensing ranges. Numerical results demonstrate the tracking accuracy and robustness of the proposed approaches. The derived UCBs constitute a tool for trustworthiness evaluation of DGP approaches and the simulation results reveal that it characterizes the presence of the target states in the error bound with $88\%$ and $42\%$ higher probability in $X$ and $Y$ coordinates, respectively, than the confidence interval-based method.
\end{abstract}


%

\section{Introduction}
Target tracking in wireless sensor networks (WSNs) is a fundamental task for various applications including sea surveillance, autonomous vehicles, and traffic monitoring. The objective is to collect sensor measurements from one or multiple targets to estimate their current and future states\cite{souza2016target}. However, the measurements may not only originate from the targets but also from the environmental interference (e.g. from ground and rain), which is referred to as the clutter~\cite{MIHAYLOVA20141}. {\color{black}To achieve reliable performance, a tracker should be able to distinguish between target measurements and clutter measurements, and also decide which measurement is associated with which target, which is called the data association problem \cite{bar2011tracking}.} 


To improve the tracking performance, numerous model-based approaches have been proposed including Kalman filter, extended Kalman filter~\cite{anderson2012optimal}, unscented Kalman filter~\cite{julier2004unscented}, and particle filter~\cite{6494261,6644279}. {\color{black}The posterior Cram\'er Rao Lower bound (PCRLB) for tracking can be calculated with different methods, e.g. as it calculated in \cite{1561880,ristic2003beyond,8455619,8882261,4350309}.}

However, these approaches rely on well-defined motion models, in particular, the target dynamics model and the sensor measurement model, which can be inaccurate when the target undergoes non-stationary evolution or mixed maneuvering behaviours. The multiple-model method~\cite{1561886} can capture complex behaviours by running a bank of elemental filters, each based on a unique model in the set and generating the overall estimates based on the results of these elemental filters. On the downside, this framework suffers from high computational complexity, and therefore, is not efficient when a large number of models are involved. 

Tracking with multiple models can be achieved by Gaussian process (GP)-based model-free methods, which are powerful non-parametric machine learning inference methods~\cite{RasmussenW06}. Instead of tracking via motion models, GP-based methods can directly learn unknown functions, which can be considered as a mapping from some system context to the target state, from noisy measurements. Particularly, GP regression can provide uncertainty quantification on the predictions. Ignoring this uncertainty can have disastrous consequences, especially when the output of such models is then fed into higher-level decision-making procedures. Recently, GP regression has been applied to solve both point target tracking~\cite{9011310,9185002,9190413,9272174,9343278,9841315,9841257} and extended target tracking~\cite{8601344,8786130,9387269} problems.  

Most of the existing GP-based tracking methods assume that the sensor measurements are collected in a centralized manner and both training and state estimation are made upon the aggregated measurements \cite{8455789,7021969}. However, due to the nature of distributed sensing systems, collecting all the measurements for GP training can bring a high communication cost, which is energy inefficient. In addition, considering a WSN with sheer amounts of sensors and measurements, the centralized tracking framework has inevitably reached an inherent bottleneck of scalability, which stems from the cubic computational complexity ($\mathcal{O}(N^3)$, where $N$ in the number of measurements) of the standard GP regression due to the inversion and determinant calculations of the GP covariance matrix. Therefore, transferring the centralized-based approaches into distributed ones has become a popular choice, and multiple approaches including distributed consensus approaches~\cite{DBLP:journals/ejasp/AliISS18} and message passing methods~\cite{meyeretal:2028:messagepassing} have been proposed for target tracking. However, to the best of the authors' knowledge, there are rarely any studies about leveraging GP regression to solve the tracking and data association in a distributed way and provide theoretical performance analysis. Hence, in this paper, the distributed Gaussian process (DGP) framework is adopted to design a DGP-based tracking (DGPT) approach, which is able to learn the target motion online and solve the data association jointly in a data-driven way. 

{\color{black}GP has also been studied in resource allocation problems under the edge computing framework to account for efficient decision-making and network state prediction \cite{mehrizi2019bayesian,da2018resource}. The presented DGPT approach can also rely on edge computing thanks to their distributed learning nature and since learning happens near the areas of data collection. The DGP framework has been implemented and tested on edge devices \cite{DA2019208}, and the whole edge learning process can be implemented in other ways as summarized in the survey~\cite{Murshed:2021:edgelearning}.}

The main contributions of this paper are summarized below:
\begin{itemize}
    \item A DGPT approach is proposed for point target tracking under the assumption that the number of measurements follows a Poisson distribution. The proposed approach can leverage both temporal and spatial features to learn the hyperparameters of the DGP online, through a sliding window of measurements.
    \item {\color{black}To solve the data association problem, different weights are assigned to the measurements based on the marginal likelihood of local GPs and a weighted summation is calculated for DGP training. This method achieves efficient hyperparameter learning and state prediction. We justify that the complexity of the DGPT does not scale with the number of measurement, but only with the length of the sliding window and the number of active sensors.}
    \item {\color{black} For the first time this work derives theoretically upper confidence bounds (UCBs) for the state estimation error of the proposed DGPT. Numerical results demonstrate the superiority of the UCBs as compared to the confidence interval of the DGP model itself.}
    \item With the knowledge learned from the DGP, a novel hybrid Bayesian filtering method is proposed to combine distributed machine learning and classical Bayesian inference. The designed tracker refines DGPT's state estimation by introducing a Poisson likelihood model, which elegantly sidesteps the data association challenge.
    \item The performance of the proposed DGP and hybrid Bayesian filtering approaches are tested with challenging target trajectory scenarios - from uniform motions to highly maneuvering ones. Different levels of measurement noise and clutter rates are involved to evaluate the accuracy and robustness of the proposed approaches.
\end{itemize}

The remaining part of the paper is organized as follows. Section II reviews the related works. Section III introduces the fundamentals of GP regression and multiple DGP methods. Section IV describes the proposed DGPT approach followed by the theoretical analysis of the tracking error bound in Section V. Section VI describes the DGP-assisted hybrid Bayesian filtering approach. The simulation setup and results are presented in Section VII, and the conclusions are drawn in Section VIII. Appendices A, B, and C contain details about the theoretical derivations.

\section{Related Work}
A wealth of approaches for improving the scalability of the standard GP method have been studied including centralized, distributed, and hybrid methods~\cite{albertandimsland:2018:survey}. In \cite{snelson2006sparse,titsias2009variational}, sparse approximations of the original $N\times{N}$ covariance matrix of GP are obtained to summarize the dependence of the whole training data using $M$ inducing points, which greatly reduces the computational complexity to $\mathcal{O}(NM^2)$. However, this type of methods requires all the data to make predictions and thus still works in a centralized manner. 

The DGP which originates from the idea of divide-and-conquer~\cite{8951257}, focuses on training local GPs (which are also referred to as local experts) based on subsets of the whole training data or based on the partitioning of the big state vector into state vectors with smaller dimensions \cite{089976602760128018}. After training the local experts, a family of aggregation methods \cite{cao2014generalized,089976600300014908,deisenroth15,cohen2020healing} relying on the product of experts model, can be applied to aggregate the local knowledge together by multiplying local predictions, and then the overall prediction can be calculated. In addition, hybrid GP variants are studied to utilize a master GP expert to communicate with the local experts leading to a consistent posterior predictive distribution \cite{liu2018generalized}. DGP has been applied to problems such as the received signal strength-based location fingerprinting map construction \cite{7870565}.

GP regression is also applied to solve the data association problem. In \cite{lazaro2012overlapping}, GP priors are placed on the different generative processes and the associations are modelled via a latent association matrix and inference is carried out using an expectation-maximization algorithm. \cite{kaiser2019data} extends GP-based data association into the non-stationary process where a different number of generative processes can be activated in different locations in the input space.
GP has been also combined with state-space models to solve the tracking problem. In \cite{5589113}, one-dimensional temporal GP regression models are reformulated as linear-Gaussian state-space models, which can be solved exactly with classical Kalman filter. The state-space model representation is also used in spatial-temporal GPs \cite{6530736} and non-Gaussian likelihood \cite{pmlr-v80-nickisch18a} to derive computationally efficient infinite-dimensional Kalman filtering and smoothing methods. There are also works studying using GP to represent the state-space model \cite{turner2010state}. GP is used to learn the whole or part of the state-space model and the learned functions can be integrated into a particle filter or extended Kalman filter \cite{8880505,9479713}, which results in hybrid tracking approach. Moreover, recently, different GP approaches are developed by assuming temporal and spatial correlation in the target trajectory and shape, and the state-space model is directly learned from the measurements \cite{9011310,9185002,8601344}. {\color{black}In Table I, we have summarized and compared more existing relevant works in GP-assisted target tracking and localization. From Table I, we can find that although GP methods have been used for solving tracking and localization problems, there are rarely works about using GP in a distributed system and integrating data association solutions into GP methods.} 

\begin{table*}[]\small\begin{center}{\color{black}
			\caption{\color{black}A summary of recent works on GP-assisted tracking and localization}
			\label{related_work_comp}
			\begin{tabular}{|c|c|c|c|c|}\hline
				\textbf{Target type}&\textbf{Ref}& \textbf{Data association}& \textbf{Centralized~/~Distributed}&\textbf{Data~/~Model-driven}\\ \hline
				\multirow{12}{*}{Point target}& \cite{9011310}& No& Centralized&Data-driven   \\ \cline{2-5} 
				& \cite{9185002}& No& Centralized&Data-driven (hybrid)\\ \cline{2-5} 
				& \cite{9272174}& No& Centralized& Data-driven (hybrid)\\ \cline{2-5} 
				& \cite{9343278}& No& Centralized& Data-driven (hybrid)\\ \cline{2-5} 
				& \cite{9841315}&No & Distributed& Date-driven \\ \cline{2-5}
				& \cite{ilic:adaptiveconsensus:2018}&No &Distributed &Model-driven \\ \cline{2-5} 
				& \cite{DBLP:journals/ejasp/AliISS18}&Yes &Distributed &Model-driven \\ \cline{2-5} 
				& \cite{7870565}&No &Distributed & Data-driven \\ \cline{2-5} 
				& \cite{jemin:2019:icassp}& No& Distributed& Data-driven\\ \cline{2-5}
				& \cite{8455789} &No &Centralized & Data-driven (hybrid) \\ \cline{2-5}
				&\cite{7021969} &Yes &Centralized & Data-driven \\ \cline{2-5}
				& \cite{RAITOHARJU2020107330}&No & Centralized& Date-driven (hybrid) \\ \cline{2-5}
				 \hline 
				\multirow{4}{*}{Group/extend target} &\cite{8601344} &No &Centralized &Data-driven \\ \cline{2-5} 
				& \cite{8786130}& Yes & Centralized& Data-driven (hybrid) \\ \cline{2-5} 
				& \cite{9387269}& No &Centralized & Data-driven \\ \cline{2-5}
				& \cite{meyeretal:2028:messagepassing}&Yes &Distributed &Model-driven \\  \hline
		\end{tabular}}
	\end{center}
\end{table*}
The next section presents background knowledge for the GP methodology before presenting the distributed learning and tracking approaches. 

\section{Background Knowledge}\label{sec:GPBT}
\subsection{Standard Gaussian Process Method}
GP is a stochastic process defining a distribution over functions that fit a set of points. Assume there exist correlations among target motions, at input $\mathbf{x}\in \mathbb{R}^d$, the non-linear mapping $f(\cdot)$ between the current input feature and the target state $f(\mathbf{x})$ can be modelled by a GP as
\begin{align}
	f(\mathbf{x})&\sim \mathcal{GP}(m(\mathbf{x}),k(\mathbf{x},\mathbf{x}')),\label{GP_model}\\
	m(\mathbf{x})&=\mathbb{E}\left[f(\mathbf{x})\right],\\
	k(\mathbf{x},\mathbf{x}')&=\mathbb{E}\left[\left(f(\mathbf{x})-\mathbf{x}\right)\left(f(\mathbf{x}')-\mathbf{x}'\right)\right],
\end{align}
where $m(\mathbf{x})$ and $k(\mathbf{x},\mathbf{x}')$ denote the mean and covariance functions, respectively. The training and the test input data are denoted by $\mathbf{x}$ and $\mathbf{x}^\prime$, respectively. Popular examples of covariance functions include the squared exponential kernel. Namely,
\begin{align}
    k(\mathbf{x},\mathbf{x}')=\sigma^2\exp\left\{-\frac{1}{2}\sum_{j=1}^d\frac{(x_j-x_j')^2}{l_j^2}\right\},
\end{align}
{\color{black}where $l_j$ represents the length-scale of the $j^\text{th}$ feature of the input data. The length-scale describes how smooth a function is which can be thought of as roughly the distance you have to move in input space before the function value can change significantly \cite{RasmussenW06}. The output variance $\sigma^2$ acts as a scaling factor. It determines the variation of function values from their mean.}

The collected measurements can be treated as the noisy outputs of the unknown functions (which are target states in the tracking problem), therefore, a point target tracking problem with noisy observations can be written as
\begin{align}\label{measure_model}
	z=f(\mathbf{x})+\epsilon,~ \epsilon \sim \mathcal{N}(0,\sigma_z^2),
\end{align}
where $z$ represents the measurement and $\epsilon$ represents the i.i.d. zero-mean Gaussian measurement noise with variance $\sigma_z^2$. 

Given a training data set of input-output pairs $D=\{\mathbf{X}, \mathbf{z}\}$ with $\mathbf{X}=\left[\mathbf{x}_1^\intercal,\mathbf{x}_2^\intercal,\cdots,\mathbf{x}_n^\intercal\right]^\intercal$ and $\mathbf{z}=\left[z_1,z_2,\cdots,z_n\right]^\intercal$, define $\mathbf{K}=k(\mathbf{X},\mathbf{X})$ as the covariance matrix of the training input, and $\mathbf{k_*}=k(\mathbf{X},\mathbf{x}_*)$ as the covariance between the training input $\mathbf{X}$ and test input $\mathbf{x}_*$. To make a prediction of the target state at a new input $\mathbf{x}_*$, the predictive mean $\mu_*$ and the predictive variance $\sigma_*^2$ can be written as
\begin{align}
	&\mu_*=m(\mathbf{x}_*)+\mathbf{k}_*^\intercal \mathbf{\Sigma}^{-1}(\mathbf{z}-m(\mathbf{x}_*)),\label{predicted_mean}\\
	&\sigma_*^2=k(\mathbf{x}_*,\mathbf{x}_*)-\mathbf{k}_*^\intercal\mathbf{\Sigma}^{-1}\mathbf{k_*},\label{predicted_var}
\end{align}
where $\mathbf{\Sigma}=\mathbf{K}+\sigma_z^2\mathbf{I}$, with $\mathbf{I}$ being the identity matrix.



\vspace{-2mm}
\subsection{Distributed Gaussian Process Method}\label{DGP}
The computational complexity and storage cost are major challenges for the large-scale learning problems. The computations require $\mathcal{O}(N^3)$ time with a standard GP implementation, where $N$ represents the number of the training instances. Besides, the standard GP also requires $\mathcal{O}(N^2+Nd)$ of memory, where $d$ is the dimensionality of
the data. Both facts limit the scalability of the standard GP regression. Moreover, according to \eqref{predicted_mean} and \eqref{predicted_var}, the standard GP can only make predictions based on all the available data, which is a centralized scheme and requires data to be shared among local GPs.

In this section, inspired by the idea of divide-and-conquer, DGP methods are introduced to reduce not only the computational cost but also the memory cost of the standard GP, by first training local GPs based on subsets of the whole training data set and then aggregating the knowledge of local GPS to achieve more accurate high-level predictions. The overall computational complexity and the memory cost can be reduced to $\mathcal{O}(Nn^2)$ and $\mathcal{O}(Mn^2+Nd)$ ($n\ll N$), respectively, where $M$ represents the number of local GPs and $n$ represents the size of data used for training a local GP. Particularly, both the computational complexity and storage cost can be further reduced through parallel/distributed computing \cite{gramacy2016lagp}.

The first type of DGP methods is the product of experts (PoEs) \cite{089976602760128018} approach. The idea is to multiply the local predictive probability distributions for overall predictions. Given the data $D^{(i)}$ collected by sensor $i$, the PoE predicts a function value $f(\mathbf{x}_*)$ at a corresponding test input $\mathbf{x}_*$ according to
\begin{align}\label{PoE}
	p(f(\mathbf{x}_*)\mid \mathbf{x}_*,D)=\prod\nolimits_{i=1}^{M}p_i(f(\mathbf{x}_*)\mid \mathbf{x}_*,D^{(i)}),
\end{align}
where $M$ is the number of GP experts and represents the number of active sensors which have measurements. Since the product of these Gaussian predictions is proportional to a Gaussian distribution, the aggregated predictive mean and variance can be calculated with closed form as
\begin{align}
	\mu^\text{PoE}_*&=(\sigma^\text{PoE}_*)^2\sum^M_{i=1}\sigma^{-2}_i(\mathbf{x}_*)\mu_i(\mathbf{x}_*),\label{poe1}\\
	(\sigma^\text{PoE}_*)^{-2}&=\sum^M_{i=1}\sigma^{-2}_i(\mathbf{x}_*),\label{poe2}
\end{align}
where $\mu_i(\mathbf{x}_*)$ and $\sigma^2_i(\mathbf{x}_*)$ represent the predictive mean and variance of GP expert $i$, respectively, which can be calculated based on $\eqref{predicted_mean}$ and $\eqref{predicted_var}$.

The PoE model provides a straightforward way to aggregate local predictions and sidesteps the weight assignment issue in other DGP models such as the mixture of expert model \cite{6215056}. However, this model becomes overconfident when making predictions, especially in regions without any training data. 

The generalized product of experts (GPoEs) model \cite{cao2014generalized} improves PoE by adding weights that represent the contributions of different experts. For instance, the weight can be calculated as the difference in the differential entropy between the prior distribution $p(f(\mathbf{x_*}))$ and the posterior predictive distribution $p(f(\mathbf{x}_*)\mid \mathbf{x}_*,D)$, which can be written as
\begin{align}
    \beta_i=0.5\left(\log\sigma^2_{**}-\log\sigma^2_i(\mathbf{x}_*)\right),
\end{align} 
where $\sigma^{2}_{**}$ represents the variance of the prior distribution $p(f(\mathbf{x}_*))$ and $\sigma^2_i(\mathbf{x}_*)$ denote the predictive variance of GP expert $i$, which can be calculated based on $\eqref{predicted_var}$.

Given the data $D^{(i)}$ collected by sensor $i$, the GPoE predicts a function value $f(\mathbf{x}_*)$ at a test input $\mathbf{x}_*$. The predictive distribution and the closed forms of the aggregated predictive mean and variance can be written as
\begin{align}
	&p(f(\mathbf{x}_*)\mid \mathbf{x}_*,D)=\prod_{i=1}^{M}p_i^{\beta_i}(f(\mathbf{x}_*)\mid \mathbf{x}_*,D^{(i)}),\\
	&\mu^\text{GPoE}_*=(\sigma^\text{GPoE}_*)^2\sum^M_{i=1}\beta_i\sigma^{-2}_i(\mathbf{x}_*)\mu_i(\mathbf{x}_*),\label{gpoe1}\\
	&(\sigma^\text{GPoE}_*)^{-2}=\sum^M_{i=1}\beta_i\sigma^{-2}_i(\mathbf{x}_*).\label{gpoe2}
\end{align}

Alternatively, the Bayesian committee machine (BCM) \cite{089976600300014908} proposes to aggregate the experts’ predictions from another view by imposing a conditional independence assumption that $D^{(i)} \indep D^{(j)}\mid f(\mathbf{x}_*)$ which in turn explicitly introduces a common prior $p(f(\mathbf{x}_*)\mid \mathbf{x}_*)$ for the experts. Therefore, the BCM's prediction distribution can be written as 
\begin{align}
	p(f(\mathbf{x}_*)\mid \mathbf{x}_*,D)=\frac{\prod_{i=1}^{M}p_i(f(\mathbf{x}_*)\mid \mathbf{x}_*,D^{(i)})}{p^{M-1}(f(\mathbf{x}_*)\mid \mathbf{x}_*)},
\end{align}
where the denominator reaches an $(M-1)$-fold division by the prior, which plays the role of a correction term that helps to recover the GP prior when leaving regions of training data. The closed form of the aggregated predictive mean and variance of the BCM can be calculated as
\begin{align}
	\mu^\text{BCM}_*&=(\sigma^\text{BCM}_*)^2\sum^M_{i=1}\sigma^{-2}_i(\mathbf{x}_*)\mu_i(\mathbf{x}_*),\label{bcm1}\\
	(\sigma^\text{BCM}_*)^{-2}&=\sum^M_{i=1}\sigma^{-2}_i(\mathbf{x}_*)+(1-M)\sigma^{-2}_{**}.\label{bcm2}
\end{align}

In~\cite{deisenroth15}, a robust Bayesian committee machine (RBCM) is proposed which combines both the features of the GPoE and BCM models. The predictive distribution and aggregated predictive mean and variance of the RBCM can be written as
\begin{align}
	&p(f(\mathbf{x}_*)\mid \mathbf{x}_*,D)=\frac{\prod\nolimits_{i=1}^{M}p_i^{\beta_i}(f(\mathbf{x}_*)\mid \mathbf{x}_*,D^{(i)})}{p^{\sum\nolimits_i\beta_i-1}(f(\mathbf{x}_*)\mid \mathbf{x}_*)},\label{RBCM1}\\
	&\mu^\text{RBCM}_*=(\sigma^\text{RBCM}_*)^2\sum\nolimits^M_{i=1}\beta_i\sigma^{-2}_i(\mathbf{x}_*)\mu_i(\mathbf{x}_*),\label{RBCM2}\\
	&(\sigma^\text{RBCM}_*)^{-2}=\sum\nolimits^M_{i=1}\beta_i\sigma^{-2}_i(\mathbf{x}_*)+(1\!\!-\!\!\sum\nolimits_{i=1}^M\beta_i)\sigma^{-2}_{**}.\label{RBCM3}
\end{align}

All the models discussed in this section can be applied to infer the target states in a distributed way in the target tracking problem. Particularly, the closed-form of posterior predictions can be obtained and the predictions are fully tractable. 

\subsection{Hyperparameter Learning}
The hyperparameters of GP need to be learned from the data. As a standard GP, maximum likelihood estimation (MLE) is applied to learn the hyperparameters by maximising the log marginal likelihood which can be written as
\begin{align}\label{mlf}
	\log p(\mathbf{z}|\mathbf{X},\pmb{\theta})&=-\frac{1}{2}\mathbf{z}^\intercal\mathbf{\Sigma}^{-1}\mathbf{z}-\frac{1}{2}\log \lvert \mathbf{\Sigma} \rvert -\frac{n}{2}\log 2\pi,
\end{align}
where $\pmb{\theta}=\left\{\sigma^2,\sigma_z^2, l_1,\cdots l_d\right\}$ represents the set of hyperparameters. {\color{black}$l_j$ represents the length-scale of the $j^\text{th}$ feature of the input data, $\sigma^2$ is the output variance of the kernel function, and $\sigma_z^2$ is the variance of the measurement noise.}
%

For the DGP, assuming the local GPs are independent with each other, the log marginal likelihood can be factorized as
\begin{align}
	&~\log p(\mathbf{z}|\mathbf{X},\pmb{\theta})\nonumber\\\approx&~ \sum\nolimits_{i=1}^{M} \log p_i(\mathbf{z}^{(i)}|\mathbf{X}^{(i)},\pmb{\theta}),\nonumber\\
	=&~\sum\nolimits_{i=1}^{M}\left( -\frac{1}{2}{\mathbf{z}^{(i)}}^\intercal{\mathbf{\Sigma}^{(i)}}^{-1}{\mathbf{z}^{(i)}}-\frac{1}{2}\log \lvert {\mathbf{\Sigma}^{(i)}} \rvert -\frac{n}{2}\log 2\pi\right),\label{a_mlf}
\end{align}
where $\mathbf{X}^{(i)}$ and $\mathbf{z}^{(i)}$ represent the training input and output of local GP $i$, respectively.

{\color{black}The reduction in computational complexity of DGP can also be justified by looking into (22), where both the computations of determinant and inversion are only based on a much smaller matrix $\mathbf{\Sigma}^{(i)}$. In addition, as compared to \eqref{mlf}, the factorized marginal likelihood can potentially be maximized in a decentralized manner like federated learning \cite{9718315} since it is a summation over local marginal likelihood functions. 
	
The learned hyperparameters are shared by all the local experts for automatic regularization to avoid overfitting.}

\vspace{4mm}\section{DGP-based Point Target Tracking}
{\color{black}The previous section demonstrates that the DGP is a promising method for large-scale learning systems. In this section, we describe the proposed DGPT approach which solves the tracking problem in WSNs using the distributed machine learning method, in a data-driven way. Several improvement schemes are designed to help integrate DGP for efficient distributed tracking and deal with clutter measurements to achieve robust performance.

In a WSN, each sensor can collect its own measurements and the GP regression can be applied locally to process local data. Some sensors can be edge devices and could provide edge learning~\cite{Murshed:2021:edgelearning}. The proposed DGPT is linked with federated learning~\cite{gafnietal:2022:federatedlearning}.

At each time, after training, local GP-based target state estimations can be aggregated to reach a high-level prediction following different aggregation methods discussed in Section \ref{DGP}. Only the the predictive means and variances of local GPs are propagated to calculate the overall prediction, without transmitting all the data to a central filter (controller). Particularly, having this aggregation process does not mean an extra central node is necessary. The aggregation can be implemented on any capable sensor or edge node, thus the proposed approach is fully distributed.}


\begin{figure}[t]
    \centering
	\includegraphics[width=7.5cm]{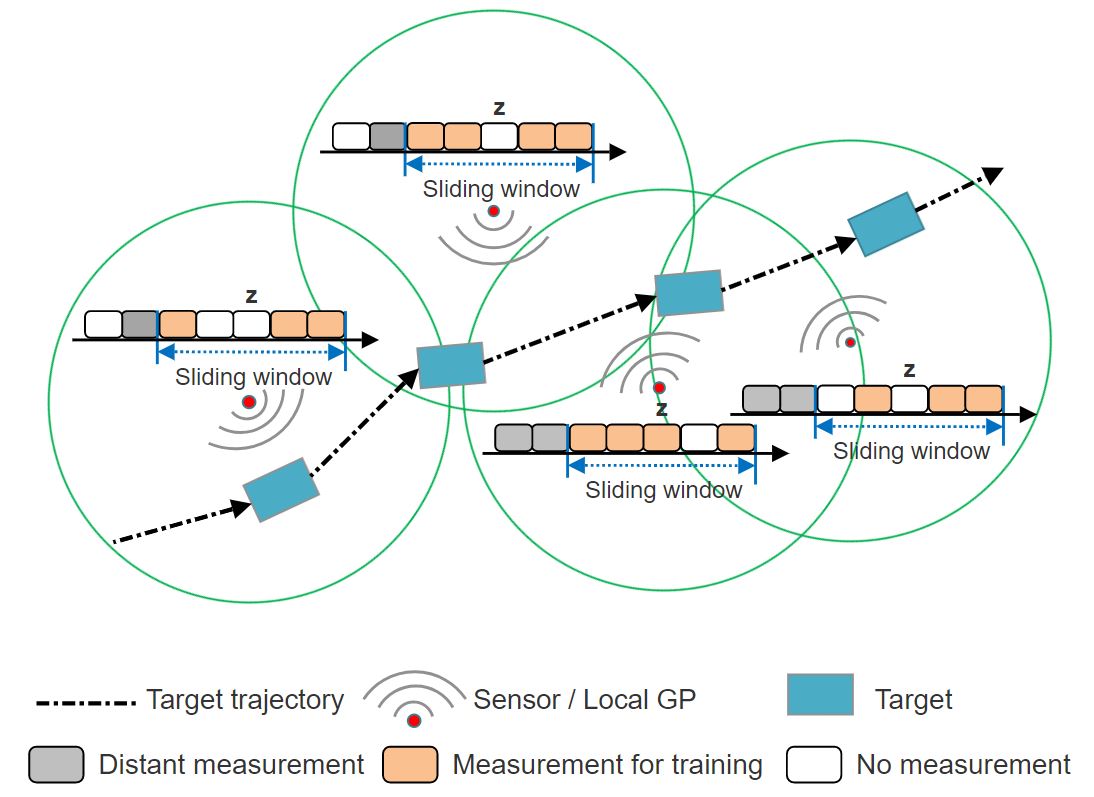}
	\caption{A distributed point tracking system with 4 sensors. The length of sliding window in this example is 5 time steps}\label{frame}
\end{figure}
\subsection{Temporal and Spatial-Temporal GP}\label{sec:4.1}
To make a state estimation of a target, it is reasonable to assume that there is a temporal correlation in the motions of the target, and this correlation within time variables in distant past is weaker than in more recent ones. Therefore, the target state can be formulated as a function of the time variable which is used as the input data for training DGP and making state estimations. In this case, we have $\mathbf{x}=t$, where $t$ represents the time variable. Based on the temporal GP (TGP), the tracking problem in \eqref{GP_model} and \eqref{measure_model} can be reformulated as
\begin{align}
    &f(t)\sim \mathcal{GP}(m(t),k(t,t')),\\
    &z=f(t)+\epsilon,~ \epsilon \sim \mathcal{N}(0,\sigma_z^2),
\end{align}
where $t$ and $t^\prime$ are the training and test data, respectively. 

{\color{black}Based on the time variables and measurements (training set), set the test input as $\mathbf{x_*}=t$, TGP can predict the target state at time $t$ using \eqref{predicted_mean} and \eqref{predicted_var}. Moreover, TGP can also predict a next-step state predictions by setting $\mathbf{x_*}=t+1$.}

Inspired by \cite{BERNTORP2021109613} that uses GP regression to learn the target state transition function, the spatial correlation can also be involved for state prediction by including the target state of the previous time into the input data, namely $\mathbf{x}=(r_{t-1},t)$, where $r_{t-1}$ represents the target state in $t-1$. The spatial-temporal GP (STGP)-based tracking problem can be formulated as
\begin{align}
    &f(r_{t-1},t)\sim \mathcal{GP}(m(r_{t-1},t),k(r_{t-1},t; r_{t-1}',t')),\\
    &z=f(r_{t-1},t)+\epsilon,~ \epsilon \sim \mathcal{N}(0,\sigma_z^2).
\end{align}
\vspace{-6mm}
\subsection{Sliding Window-based Tracking}
Although DGP is designed for complexity reduction, training GPs distributively can still be computationally intense when local sensors collect extensive measurements originating from both the targets and the clutter. 

Noticing the fact that the target motion can be a mixed maneuvering behaviour with time-varying parameters, based on the temporal correlation assumptions in the previous section, measurements with weaker motion correlations in distant past time cannot contribute to training and prediction as much as the recent ones. Therefore to further reduce the computational complexity, the distant measurements are abandoned in DGPT and only the recent measurements are utilized for training, which can be treated as using a sliding window to select valid measurements. The sliding window is a set of time variables which represents a number of time steps. Based on the sliding window, a sensor without valid measurements (the times of collecting the measurements are not in the sliding window) is excluded from state estimation at the current time. 

When both the number of experts and target measurements is reduced, the computational complexity of GP training reduces and also the tracking accuracy improves. Figure \ref{frame} shows the framework of the proposed sliding window-based DGPT approach.


\subsection{DGP-based Data Association}\label{sec:4.3}
The previous section discuss using sliding window to reduce the number of measurements for training and state estimation. In this section, we focus on dealing with a large number of clutter measurements to improve the tracking accuracy and also reduce the computational complexity.



{\color{black}To cope with the clutter and for learning the GP hyperparameters, a method is designed to assign weights to different measurements based on the marginal likelihood \eqref{mlf}.} A weighted summation over measurements collected by one sensor is then calculated as the training data for DGP. Define $j\in\{1,2,\cdots,J_t\}$ as the index of measurements received by a sensor at time $t$ and define $w_{j,t}$ as the weight of the $j^{\text{th}}$ measurement, the gating method can be written as
\begin{align}
    &w_{j,t}=p({z_j}|\mathbf{X}_t,\pmb{\theta}_{t-1})/{\sum\nolimits_j^{J_t} p({z_j}|\mathbf{X}_t,\pmb{\theta}_{t-1})},\label{gating1}\\
    &\bar{z}_t=\sum\nolimits_j^{J_t} w_{j,t} z_j,\label{gating2}
\end{align}
where $\bar{z}_t$ represents the weighted summation of the measurements, which will be used for hyperparameter learning. The rationale for calculating this weighted summation is to determine a training instance based on the likelihoods of all the measurements (both target and clutter measurements) and the resulting summation is expected to be close to the target measurements. Particularly, since only one summation is calculated from each local expert, the number of measurements for DGP training is greatly reduced and therefore the computational cost is reduced as well.

In the distributed tracking scenario, at the beginning of each time, each sensor performs the proposed gating method independently based on the learned hyperparameters from the previous time and the local measurements which are inside the sliding window. The resulting data based on \eqref{gating1} and \eqref{gating2} is used for hyperparameter learning and state estimation in the current time. In addition, for a new sensor that does not have any historical measurements, average weights are calculated according to the weights assigned by all the other active sensors. Moreover, when the clutter rate is much higher, the clustering scheme can be used for preprocessing and the gating method will be applied to the cluster centers rather than to all the measurements.


\subsection{Hyperparameter Online Learning in DGPT}
Since the target motion can be time-varying and the sliding window is designed to keep valid measurements for training and tracking, the hyperparameters of DGP should be learned online to capture the non-stationary features. Hence, in the proposed DGPT approach, MLE which is based on the factorized marginal likelihood $\eqref{a_mlf}$ is solved every time to update the hyperparameters. This process brings extra computational costs due to the non-convexity of $\eqref{a_mlf}$ and requires an iterative solving process. To accelerate the hyperparameter learning process, optimized hyperparameters at time $t$ is designed to be set as the initial value of hyperparameters of MLE at $t+1$, which can significantly reduce the iterations needed for MLE. 

Benefiting from the proposed sliding window design, the DGP-based data association, and the online learning properties, the DGPT can provide accurate target state estimations with collected measurements.

{\color{black}\subsection{Complexity Analysis}
For the proposed DGPT approach, the main computational complexity stems from the covariance matrix inversion, which scales cubically in terms of the number of data at each sensor. Based on [23], the complexity of matrix inversion for DGP is $\mathcal{O}(J_{1,t}^3+J_{2,t}^3+\cdots J_{M_t,t}^3)$, where $J_{M_t,t}$ represents the number of data stored in the ${M_t}^{\text{th}}$ active sensor at time $t$ and $M_t$ is the number of active sensors at $t$. Based on (27) and (28), the measurements collected by a sensor at time $t$ are used to calculate a weighted sum for DGP training. This means only a single measurement is saved per sensor per time. Given the length of the sliding window of time as $C$, the computational complexity due to the matrix inversion can be upper bounded as $\mathcal{O}(C^3M_t)$.

In the DGPT approach, the hyperparameters are learned online by solving the maximum likelihood estimation problem as formulated in (22). Maximizing this function requires computing the covariance matrix inversion iteratively. Therefore, the computational complexity of the DGP algorithm update scales as $\mathcal{O}(C^3M_t)$ per iteration at time $t$. The prediction step also scales as $\mathcal{O}(C^3M_t)$. It is important to notice that the proposed DGP-based data association helps the DGPT achieve a low computational complexity in both DGP learning and state prediction since the complexity does not scale with the number of received measurements and the length of the sliding window does not increase over time.}

\section{Theoretical Performance Analysis}
\begin{Lemma}\label{lemma1}(Lemma 5.1 of \cite{105555})
Given a trained local GP based on training data $D_t=\left\{\mathbf{X}_t,\mathbf{z}_t\right\}$ till $t$, for any input $\mathbf{x}_*\in \mathbf{X}_t$, the probability $\text{Pr}(.)$ that the predictive mean $\mu(\mathbf{x}_*)$ deviates from the true function value by more than a certain amount can be upper bounded as
\begin{align}
    \text{Pr}\left\{\lvert f(\mathbf{x}_*)-\mu(\mathbf{x}_*) \rvert >\gamma^{1/2}\sigma(\mathbf{x}_*)\right\}\leq e^{-\gamma/2},
\end{align}
where $\gamma$ is a positive constant.
\end{Lemma}

Lemma 1 proposes a UCB of the probability that the deviation between the true function and the estimated mean of the function is larger then a scaled version of the estimated variance function. Based on this lemma, error bounds of distributed GPs can be derived.

The following Lemma 2 represents a generalization of Lemma 1, assume each time one estimation is made, for the case of an infinite number of time $t$, e.g. $t\rightarrow \infty$.
\begin{Lemma}\label{lemma2}
Define $\delta\in(0,1)$, set $\gamma_t=2\log(\pi_t/\delta)$, for $\pi_t=\pi^2t^2/6$, based on Lemma \ref{lemma1}, apply the union bound over $t$, we have
\begin{align}
    &\text{Pr}\left\{\bigcup\nolimits_{t=1}^{\infty}\lvert f(\mathbf{x}_t)-\mu(\mathbf{x}_t) \rvert >\gamma_t^{1/2}\sigma(\mathbf{x}_t)\right\}\nonumber\\\leq& \sum\nolimits_{t=1}^\infty e^{-\gamma_t/2}
    =\sum\nolimits_{t=1}^\infty \frac{\delta}{\pi_t}=\delta.
\end{align}
\end{Lemma}

Lemma 2 further generalises the cumulative deviation of the predictive mean from the true function value based on the GP predictions from all $n$ test inputs.
\vspace{-2mm}
\begin{Theorem}
(One-step error bound of GPoE) Consider a distributed GP system with $M$ local GPs, with probability at least $1-\sum_{i=1}^M e^{-\gamma_i/2}$, the deviation between the true function value at $\mathbf{x}_*$ and the aggregated estimation of the mean value made by the GPoE method can be upper bounded as\vspace{-6mm}
\end{Theorem}
\begin{align}\label{ucb:PoE}
    \lvert f(\mathbf{x}_*)-\mu^\text{GPoE}_* \rvert&\leq \frac{\sum^M_{i=1}\gamma_i^{1/2}\sigma^{-1}_i(\mathbf{x}_*)}{\sum^M_{i=1}\sigma^{-2}_i(\mathbf{x}_*)}.
\end{align}

\textit{Proof:} Define $A_i$ as the event in which the prediction of the target state from local expert $i$ and the true target state differs larger than a quantity, which can be written as
\begin{align}
    A_i=\left\{\lvert f(\mathbf{x}_*)-\mu_i(\mathbf{x}_*) \rvert >\gamma_i^{1/2}\sigma_i(\mathbf{x}_*)\right\}.
\end{align}

Define the union of events $\left\{A_1,A_2,\cdots,A_M\right\}$ as $A$. Applying the union bounds over $M$ events, the probability of $A$ can be upper bounded as
\begin{align}
    \text{Pr}(A)&\defeq\text{Pr}\left\{ \bigcup\nolimits_{i=1}^{M} A_i \right\}\nonumber,\\&=\text{Pr}\left\{ \bigcup\nolimits_{i=1}^{M} \lvert f(\mathbf{x}_*)-\mu_i(\mathbf{x}_*) \rvert >\gamma_i^{1/2}\sigma_i(\mathbf{x}_*) \right\}\nonumber,\\&\leq \sum_{i=1}^M e^{-\gamma_i/2},
\end{align}
where $\mu_i(\mathbf{x}_*)$ and $\sigma_i(\mathbf{x}_*)$ represent the predictive mean and standard deviation (STD) of local GP $i$ at $\mathbf{x}_*$, respectively.

Define $\Bar{A}$ as the complement of A, changing the direction of the inequality gives that
\begin{align}\label{bound1}
     \text{Pr}(\Bar{A})&=\text{Pr}\left\{ \bigcap\nolimits_{i=1}^{M} \Bar{A}_i \right\}\nonumber,\\&=\text{Pr}\left\{ \bigcap\nolimits_{i=1}^{M} \lvert f(\mathbf{x}_*)-\mu_i(\mathbf{x}_*) \rvert \leq\gamma_i^{1/2}\sigma_i(\mathbf{x}_*) \right\},\nonumber\\&\geq 1-\sum\nolimits_{i=1}^M e^{-\gamma_i/2}.
\end{align}

According to~\eqref{gpoe1} and \eqref{gpoe2}, the deviation between the true function value and the aggregated predictive mean by GPoE can be written as
\begin{align}\label{bound2}
    \bigg\lvert f(\mathbf{x}_*)-\mu^\text{GPoE}_* \bigg\rvert&=\bigg\lvert f(\mathbf{x}_*)-\frac{\sum^M_{i=1}\beta_i\sigma^{-2}_i(\mathbf{x}_*)\mu_i(\mathbf{x}_*)}{\sum^M_{i=1}\beta_i\sigma^{-2}_i(\mathbf{x}_*)}\bigg\rvert\nonumber,\\
    &=\frac{\lvert\sum^M_{i=1}\beta_i\sigma^{-2}_i(\mathbf{x}_*) \left(f(\mathbf{x}_*)-\mu_i(\mathbf{x}_*)\right)\rvert}{\sum^M_{i=1}\beta_i\sigma^{-2}_i(\mathbf{x}_*)}\nonumber,\\
    &\leq\frac{\sum^M_{i=1}\beta_i\sigma^{-2}_i(\mathbf{x}_*)\lvert f(\mathbf{x}_*)-\mu_i(\mathbf{x}_*)\rvert}{\sum^M_{i=1}\beta_i\sigma^{-2}_i(\mathbf{x}_*)}.
\end{align}

Based on \eqref{bound1}, with probability at least $1-\sum_{i=1}^M e^{-\gamma_i/2}$, \eqref{bound2} can be upper bounded as
\begin{align}
    \lvert f(\mathbf{x}_*)-\mu^\text{GPoE}_* \rvert&\leq \frac{\sum^M_{i=1}\beta_i\sigma^{-2}_i(\mathbf{x}_*)\gamma_i^{1/2}\sigma_i(\mathbf{x}_*)}{\sum^M_{i=1}\beta_i\sigma^{-2}_i(\mathbf{x}_*)},\nonumber\\
    &=\frac{\sum^M_{i=1}\beta_i\gamma_i^{1/2}\sigma^{-1}_i(\mathbf{x}_*)}{\sum^M_{i=1}\beta_i\sigma^{-2}_i(\mathbf{x}_*)},
\end{align}
which completes the proof.

Theorem 1 proposes a theoretical UCB for the tracking performance. Define the highest predictive variances of a local expert as $\sigma^2_\text{H}$, the bound can be further represented as
\begin{align*}
    &1-\sum\nolimits_{i=1}^M e^{-\gamma_i/2}\nonumber\\\leq&
     \text{Pr}\left\{\lvert f(\mathbf{x}_*)-\mu^\text{GPoE}_* \rvert\leq \frac{\sum^M_{i=1}\beta_i\gamma_i^{1/2}\sigma_i^{-1}(\mathbf{x}_*)}{\sum^M_{i=1}\beta_i\sigma^{-2}_i(\mathbf{x}_*)}\right\},\nonumber\\
    \leq&\text{Pr}\left\{\lvert f(\mathbf{x}_*)-\mu^\text{GPoE}_* \rvert\leq \frac{\sum^M_{i=1}\beta_i\gamma_i^{1/2}\sigma_i^{-1}(\mathbf{x}_*)}{M\sigma^{-2}_\text{H}(\mathbf{x}_*)\sum^M_{i=1}\beta_i}\right\}.\nonumber\\
\end{align*}
This bound demonstrates that, given all other local GPs fixed, when one of the local GP makes a highly uncertain prediction which is reflected as a larger predictive variance, the upper bound of the deviation will increase, which means the overall prediction is exacerbated by this poor GP expert.

Next, we derive a Theorem about the UCB of the RBCM. 

\begin{Theorem}
(One-step error bound RBCM) following Theorem 1, for a distributed GP system with $M$ local GPs, with probability at least $1-\sum_{i=1}^M e^{-\gamma_i/2}$, the deviation between the true function value at $\mathbf{x}_*$ and the aggregated estimation of the mean value made by the RBCM method can be upper bounded as\vspace{-6mm}
\end{Theorem}
\begin{align}\label{ucb:BCM}
    \lvert f(\mathbf{x}_*)-\mu^\text{RBCM}_* \rvert&\leq \frac{\sum^M_{i=1}\gamma_i^{1/2}\sigma^{-1}_i(\mathbf{x}_*)}{\sum^M_{i=1}\sigma^{-2}_i(\mathbf{x}_*)}.
\end{align}
\textit{Proof:} The detailed proof is given in Appendix A.

The UCB for the GPoE algorithm in general is not the same as the UCB~(\ref{ucb:PoE}) for the RBCM~(\ref{ucb:BCM}). However, under certain conditions the two upper bounds coincide. 



The next Theorem 3 generalises the result further, to the GPoE's cumulative error bound for a number of estimations over a wide time interval $T$. The cumulative error bound of other DGPT approaches can be derived in a similar way.

\begin{Theorem}
(Cumulative error bound) Consider a distributed GP system with $M$ local GPs. suppose one state estimation is made every time, with probability at least $1-\sum_{i=1}^M \delta_i$, the cumulative deviation between the true function value at each test input and the aggregated estimation of the mean value can be upper bounded as
\begin{align}
    \sum_{t=1}^T\lvert f(\mathbf{x}_t)-\mu^{\text{GPoE}}(\mathbf{x}_t) \rvert&\leq \sum_{t=1}^T\frac{\sum^M_{i=1}\beta_i\gamma_{t,i}^{1/2}\sigma^{-1}_{i}(\mathbf{x}_t)}{\sum^M_{i=1}\beta_i\sigma^{-2}_{i}(\mathbf{x}_t)}.
\end{align}
\end{Theorem}
\textit{Proof:} The detailed proof is given in Appendix B.

\section{DGP-assisted Bayesian Filtering with Measurement Origin Uncertainty}\label{sec:Likelihood_wuth_clutter_measurements}
Inspired by ideas from~\cite{Gilholm1,gilholm2005spatial,phdfilterfortraffic}, in this section, the proposed DGPT is enhanced by an elegant Bayesian filtering method which can solve the data association problem without the need to construct explicit measurement-target assignment hypotheses or gates. The resultant hybrid Bayesian filtering-based tracking approach provides a novel way to merge distributed machine learning and model-based Bayesian inference. The prediction made from the DGPT is used as the prior distribution of the target state and a Poisson measurement likelihood model is involved for posterior state inference.

\subsection{Measurement Likelihood Function}
{\color{black}According to~\cite{Gilholm1}, the state vector of $L+1$ entities that needs to be estimated at time $t$ is defined as
\begin{align}
    \tilde{\mathbf{X}}_t=\left[\tilde{\mathbf{x}}_{t,\text{c}}^\intercal,\tilde{\mathbf{x}}_{t,1}^\intercal,\cdots,\tilde{\mathbf{x}}_{t,L}^\intercal \right]^\intercal,
\end{align}
where $\tilde{\mathbf{x}}_{t,\text{c}}$ represents the state of the clutter process and $L$ represents the number of targets which is assumed to be known. In a single point tracking problem, we have $L=1$ and $\tilde{\mathbf{x}}_{t,1}=\tilde{\mathbf{x}}_t$.

The derivation of the measurement likelihood relies on the following three assumptions: 
\begin{itemize}
    \itema The numbers of target originated measurements in a time scan are assumed to be Poisson distributed, with a rate~$\lambda_{\text{T}}$.
    \itemb The numbers of clutter measurements in a time scan are assumed to be Poisson distributed, with a clutter rate $\lambda_{\text{c}}$
    \itemc The clutter measurements are assumed to be uniformly distributed in the sensing space of each sensor.
\end{itemize}

{\color{black}In many cases, in a time step, a high-resolution sensor is able to generate more the one measurement from the target and also from the environmental interference. Therefore, the assumptions that the clutter rate $\lambda_\text{c}$ and target rate $\lambda_\text{T}$ of the respected measurements follow Poisson distributions are well justified. These assumptions are also used in \cite{6479228,6451107}. Then this is reflected in the first and second assumptions (A1 and A2). Assumption A3 reflects the fact that the clutter is uniformly distributed which is one of the most common cases in practice. The uniform distribution of the clutter in the areas of interest also reflects the full lack of prior knowledge about the possible locations of the environmental interference.}

For a WSN, according to above assumptions and by assuming that the target originated measurement likelihood is a product of Gaussian likelihoods, the collected measurements result from superposition of multiple target measurements with Gaussian noise and uniform clutter measurements \cite{Gilholm1}. Therefore, similarly to~\cite{gilholm2005spatial} and \cite{phdfilterfortraffic}, the joint likelihood expression of the tracking problem can be written as 
\begin{align}
	p(\mathbf{z}_t|\tilde{\mathbf{X}}_t)&=p((z_{t,1},z_{t,2},\cdots,z_{t,n_t}),n_t|\tilde{\mathbf{X}}_t),\label{joint_likelihood}
\end{align}
where $n_t$ represents the number of measurements collected from all the sensors in time $t$. $\mathbf{z}_t$ denotes the measurements collected in $t$. 

To simplify the notation we rewrite $\tilde{\mathbf{x}}_t$ as $\tilde{x}_t$ to represent the case when the target state is a scalar. 

Define $p_\text{c}(z_{t,j})$ as the clutter measurement likelihood which is a uniform distribution, and define $p(z_{t,j}|\tilde{x}_t)$ as the target measurement likelihood which is Gaussian, based on the assumptions. In addition, define $\phi$ as a partition of the measurement set, the joint likelihood \eqref{joint_likelihood} can be written as

\begin{align}
    p(\mathbf{z}_t|\tilde{\mathbf{X}}_t)&\propto\sum_{\phi}\left(\frac{\lambda_{\text{T}}}{\lambda_{\text{c}}}\right)^{n_t^\text{T}(\phi)} \prod_{j,\phi(j)\neq0}^{n_t}p(z_{t,j}|\tilde{x}_t) \prod_{j,\phi(j)=0}^{n_t} p_c(z_{t,j}),
\end{align}
where $n_t^\text{T}(\phi)$ represents the number of target measurements which is compatible with partition $\phi$. $\phi(j)=0$ corresponds to the clutter measurement and $\phi(j)\neq0$ corresponds to the target measurement. $\lambda_{\text{T}}$ is the expected number of measurements originating from the target and $\lambda_{\text{c}}$ denotes the expected number of measurements corresponding to clutter in the sensing area. According to Appendix A of \cite{phdfilterfortraffic}, the likelihood can be further represented as}
\begin{align}
    p(\mathbf{z}_t|\tilde{\mathbf{X}}_t)
	&\propto\prod\nolimits_{j=1}^{n_t}\left( \lambda_{\text{c}} p_c(z_{t,j})+\lambda_{\text{T}} p(z_{t,j}|\tilde{x}_t)  \right).\label{likelihood}
\end{align}

Based on Assumption \textbf{A3}, the clutter measurement likelihood can be written as $p_\text{c}(z_{t,j})=\frac{1}{A_{\text{sen}}}$, where $A_{\text{sen}}$ represents the sensing area and is considered to be the same for all the~sensors.

The measurement likelihood is expressed in two different ways: with the Poisson likelihood model~(\ref{likelihood}) or its equivalent form~(\ref{eq:Likelihood_with_Poisson_assumptions}) 
and with a Gaussian process as in~(\ref{Gaussian_product2}).

Assuming the measurement noise follows a zero-mean Gaussian distribution with variance $\sigma_z^2$, we have $p(z_{t,j}|\tilde{x}_t) \sim \mathcal{N}(\tilde{x}_t,\sigma_z^2)$, the measurement likelihood \eqref{likelihood} can be represented as~\begin{align}
    p(\mathbf{z}_t|\tilde{\mathbf{X}}_t)&\propto \prod_{j=1}^{n_t}\left\{\frac{\lambda_{\text{c}}}{A_{\text{sen}}}+\lambda_{\text{T}} p\left(z_{t,j}|\tilde{x}_t\right)\right\}, \label{eq:Likelihood_with_Poisson_assumptions}\\
    &\propto \prod_{j=1}^{n_t}\mathcal{N}\left(\hat{\mu}_{t,j},\hat{\sigma}_{t,j}^2\right), \label{Gaussian_product1}\\
    &\propto \mathcal{N}\left(\hat{\mu}_{t},\hat{\sigma}_{t}^2\right),\label{Gaussian_product2}
\end{align}
where $\hat{\mu}_{t,j}=\frac{\lambda_{\text{c}}}{A_{\text{sen}}}+\lambda_{\text{T}} \tilde{x}_t$ and $\hat{\sigma}_{t,j}^2=(\lambda_{\text{T}}\sigma_z)^2$. The derivation from \eqref{Gaussian_product1} to \eqref{Gaussian_product2} holds due to the fact that the product of Gaussian probability density functions is proportional to a Gaussian probability density function.

Based on \eqref{poe1} and \eqref{poe2} of product of Gaussian, the mean and the variance of \eqref{Gaussian_product2} can be calculated as
\begin{align}
    \hat{\sigma}_{t}^{2}&=\frac{1}{\sum_{j=1}^{n_t^i}\hat{\sigma}_{t,j}^{-2}}=\frac{1}{\sum_{j=1}^{n_t}\left(\lambda_{\text{T}}\sigma_z\right)^{-2}}=\frac{\left(\lambda_{\text{T}}\sigma_z\right)^{2}}{n_t}, \label{eq_likelihood_with_clutter_rate_var}\\
    \hat{\mu}_{t}&= \hat{\sigma}_{t}^{2}\sum_{j=1}^{n_t}\hat{\sigma}_{t,j}^{-2}\hat{\mu}_{t,j}
    =\hat{\sigma}_{t}^{2}\sum_{j=1}^{n_t}\left(\lambda_{\text{T}}\sigma_z\right)^{-2}\cdot\left(\frac{\lambda_{\text{c}}}{A_{\text{sen}}}+\lambda_{\text{T}} \tilde{x}_t\right),\nonumber \\
    &=\hat{\sigma}_{t}^{2}n_t\left(\lambda_{\text{T}}\sigma_z\right)^{-2}\cdot\left(\frac{\lambda_{\text{c}}}{A_{\text{sen}}}+\lambda_{\text{T}} \tilde{x}_t\right),\nonumber\\
    &=\frac{\lambda_{\text{c}}}{A_{\text{sen}}}+\lambda_{\text{T}} \tilde{x}_t.\label{eq_likelihood_with_clutter_rate_mean}
\end{align}


The expressions \eqref{eq_likelihood_with_clutter_rate_var} and \eqref{eq_likelihood_with_clutter_rate_mean} are functions of the clutter parameter (e.g. of the clutter rate $\lambda_{\text{c}}$) and of the rate $\lambda_{\text{T}}$ of the target originated measurement. The likelihood function could be expressed also as a function of the clutter density, as this is shown in~\cite{gilholm2005spatial,phdfilterfortraffic}. The clutter rate $\lambda_{\text{c}}$ and the clutter density are connected with the area of the sensor and are parameters that can be estimated by knowing the sensor area $A_\text{sen}$, as shown in~\cite{defreitasetal:2016:autonomouscrowdstracking} or by theoretical methods.

\subsection{DGP-based Posterior State Inference}
{\color{black}As discussed in Section \ref{sec:4.1}, notice that at time $t$, the proposed DGPT approach can also provide a next-step state estimation distribution. This means the DGP model can be used as a high-quality prior distribution for the state estimation in the next time. Have this prior knowledge and by combining the likelihood, a novel state estimation method can be designed. }

Define $\mathbf{Z}_{t-C:t-1}$ as the set of measurements within the sliding window from time $t-C$ to $t-1$. The length of the sliding window is $C$ time steps. Given the current time $t$, the prior distribution of the target state can be written as
\begin{align}
    p(\tilde{\mathbf{X}}_t\mid \mathbf{Z}_{t-C:t-1})\sim \mathcal{N}(\mu_{t-1}, \sigma^2_{t-1}),
\end{align}
where $\mu_{t-1}$ and $\sigma^2_{t-1}$ represent the prior mean and variance that can be calculated using the GP regression equations (4) and (5), respectively. According to Bayes rule, the posterior can be written as
\begin{align}\label{bayes-update}
    p(\tilde{\mathbf{X}}_t\mid \mathbf{Z}_{t-C:t})\propto~& p(\mathbf{z}_t|\tilde{\mathbf{X}}_t)p(\tilde{\mathbf{X}}_t|\mathbf{Z}_{t-C:t-1}).
\end{align}

Based on the prior distribution \eqref{RBCM1}, \eqref{RBCM2}, and \eqref{RBCM3}, which is learned by the DGP, considering the measurement likelihood \eqref{likelihood}, the posterior state distribution \eqref{bayes-update} can be derived. The posterior mean and variance can be written as
\begin{align}
    \mu_{t}&=\frac{\mu_{t-1}\hat{\sigma}_{t}^2+\sigma_{t-1}^2\sum_{j=1}^{n_t}(z_j\lambda_{\text{T}}-\lambda_{\text{T}}\lambda_{\text{c}}/A_\text{sen})}{\hat{\sigma}_{t}^2+n_t \sigma_{t-1}^2\lambda_{\text{T}}^2},\\
    \sigma_{t}^2&=\frac{\hat{\sigma}_{t}^2\sigma_{t-1}^2}{\hat{\sigma}_{t}^2+n_t \sigma_{t-1}^2\lambda_{\text{T}}^2}.
\end{align}
Appendix~C contains the detailed derivation of these results. 

A thorough performance validation and evaluation of the developed DGPT and hybrid Bayesian filtering approaches with the derived theoretical UCB and over several test scenarios are presented in the next section. 

\section{Performance Evaluation and Validation}
\subsection{Training Time of DGP}
To evaluate the running time for training the DGP model, in this subsection, define the input data as two-dimensional, namely $\mathbf{x}=[x_1,x_2]^\top$, we test the running time of DGP training based on data from the following function
\begin{align}
    f(\mathbf{x})=5 x_1^2+\sin(12 x_2),
\end{align}
where the measurements are generated by adding zero-mean Gaussian noise with the STD as $\sigma_z=0.5$ to this function.
\begin{figure}[t]
    \centering
	\includegraphics[width=3.5in]{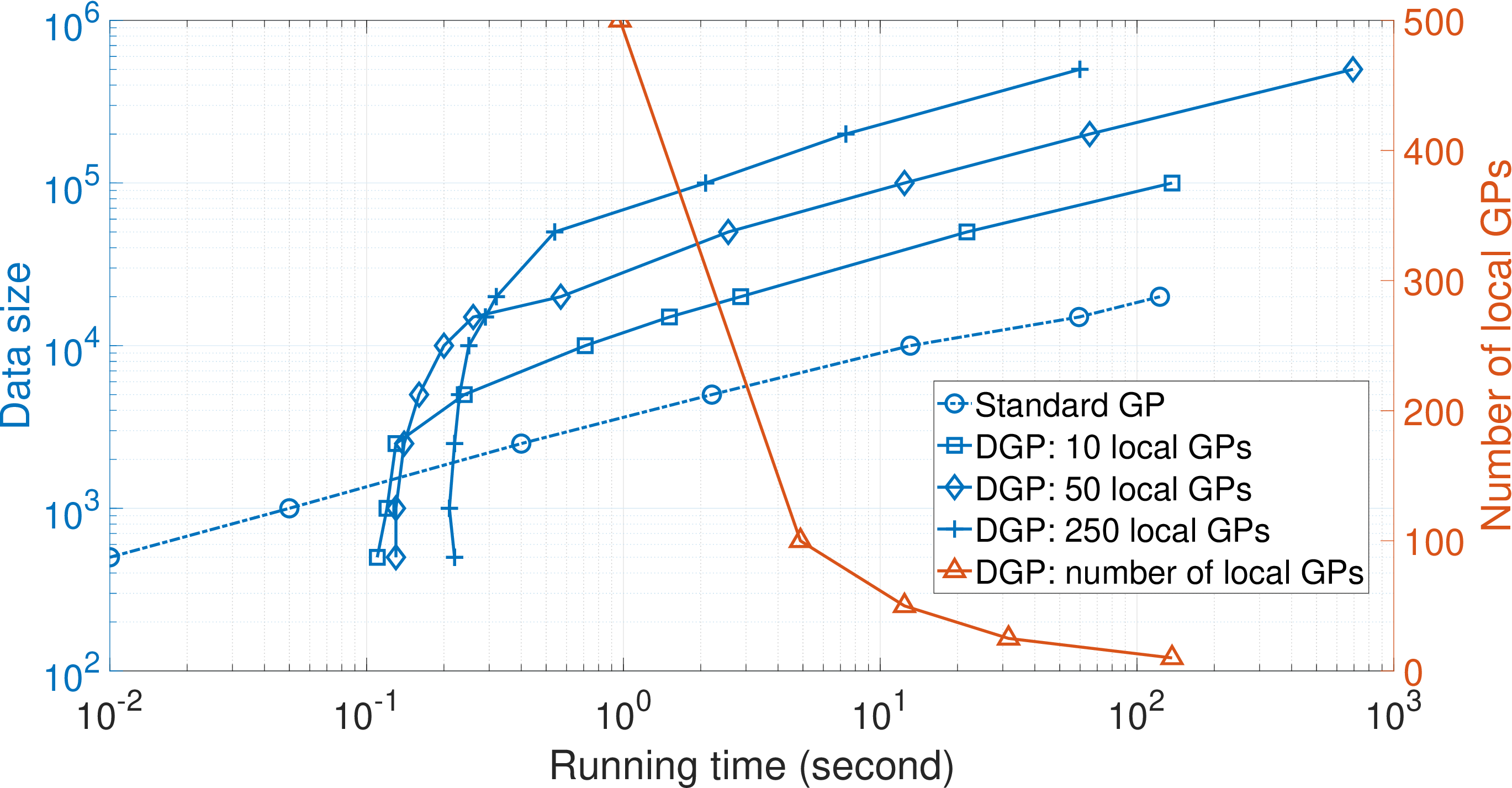}
	\caption{Computation time for the log-marginal likelihood and
its gradient versus the size of the training data and the number of local GPs}\label{time}
\end{figure}

The time required to compute the log marginal likelihood and its gradient with respect to the kernel hyperparameters is presented in Figure \ref{time}. This computation process acts as the fundamental step for solving the factorized MLE \eqref{a_mlf} for hyperparameter learning and DGP training. As a comparison, we also measure the time required for training the standard GP model, which corresponds to solving the original MLE \eqref{mlf}. In addition, we measure the running time of solving the factorized MLE with varying numbers of local GPs. According to the results, we can find that the computation time of training DGP increases with the growing size of the training data set and adding more local GPs can accelerate the training process.

Particularly, except in the cases of small training sets (500-1000), the running time of DGP increases much slower than the standard GP. Hence, DGP can handle much more data and is more suitable for real-time distributed learning and tracking in WSNs as compared to the standard GP. Moreover, the computation of the factorized likelihood can be implemented in parallel, thus each local computational unit will only need to compute one or a few terms of \eqref{a_mlf}, and the overall running time can be further reduced.

In addition, in Figure \ref{time}, the impact of the number of local GPs on the running time for computing the log marginal likelihood and its gradient is also presented based on the data set with a fixed size of 10000. We can find that having more local GPs can help reduce the running time since according to \eqref{mlf}, if the smaller size of covariance matrices are generated, solving the matrix determinant and inversion can be much faster on these smaller matrices. 
\subsection{Simulation Setup}
The tracking performance of the proposed approaches are tested in a WSN with 250 sensors uniformly implemented in a 1000 meters $\times$ 1000 meters area. The sensing range is $50$ meters and the sampling period is one second, both of which are identical for every sensor. The proposed algorithm can also be implemented in a heterogeneous network easily and can make state estimations considering the heterogeneity of sensors, which for example, can be reflected in the posterior predictive variance of the local sensor.

In this paper, the target states denotes the target locations, hence two GPs are needed at one sensor for the $X$-coordinate and $Y$-coordinate, respectively. For the GP, a zero-mean function is used which means no extra knowledge is utilized for tracking. Besides, the covariance function is selected to be the squared exponential kernel which is demonstrated to perform well in many maneuvering models \cite{9190413}. Following Assumptions A1-A3, the clutter measurements are modelled as a Poisson point process and are uniformly located in the sensing region of each active sensor. Moreover, the number of target measurements is modelled as a zero-truncated Poisson distribution, which ensures at least one measurement can be received by a sensor within a time step, namely, $\lambda_{\text{T}}=1$. The target measurement noise is modelled by the zero-mean Gaussian distribution.

In addition, we develop a range of scenarios with varying parameters of the clutter process, measurement noise, and target trajectories. {\color{black}To test the robustness of the proposed DGPT approach and also to account for the wireless effect of the channel, three noise levels are involved in the simulation with measurement uncertainty (STD) $\sigma_z=1,2,4$ meters.} In addition, two clutter settings are simulated to test the performance of the proposed approach, the low clutter case sets the clutter rate as 1 and the high clutter case sets the rate to be 5. All the results are averaged over 100 Monte Carlo (MC) runs and the lengths of sliding windows in different trajectories are carefully tuned to be different for optimal performance. 

\subsection{Benchmarks}
Since this paper focuses on the model-free approaches, the standard GP-based centralized tracking approach is simulated as the benchmark. This scheme relies on solving the MLE \eqref{mlf} to learn the hyperparameters, and the learning process requires the measurements to be transmitted in the WSN. To make fair comparisons, the standard GP-based centralized tracking approach is trained with the same sensor measurements in the sliding window, which means this approach uses the same amount of data for model training and hyperparameter learning as well as the DGPT approach.

To study the impact of different aggregation methods on the DGPT, both RBCM and GPoE are simulated. In addition, DGPTs based on both temporal and spatial-temporal input data are evaluated. For the temporal case, a set of time variables which is in the sliding window is used as the input data. For the spatial-temporal case, both the current time and the target state of the previous time are used as input. Notice that in the online tracking problem, the real target state is not available to be used as the training input. Therefore, the predictive state acquired by the DGPT in the previous time is used instead.
\subsection{Target Trajectories}
\begin{figure}[t]
\begin{minipage}[b]{.49\linewidth}
	\centering
	\centerline{\includegraphics[width=4.35cm]{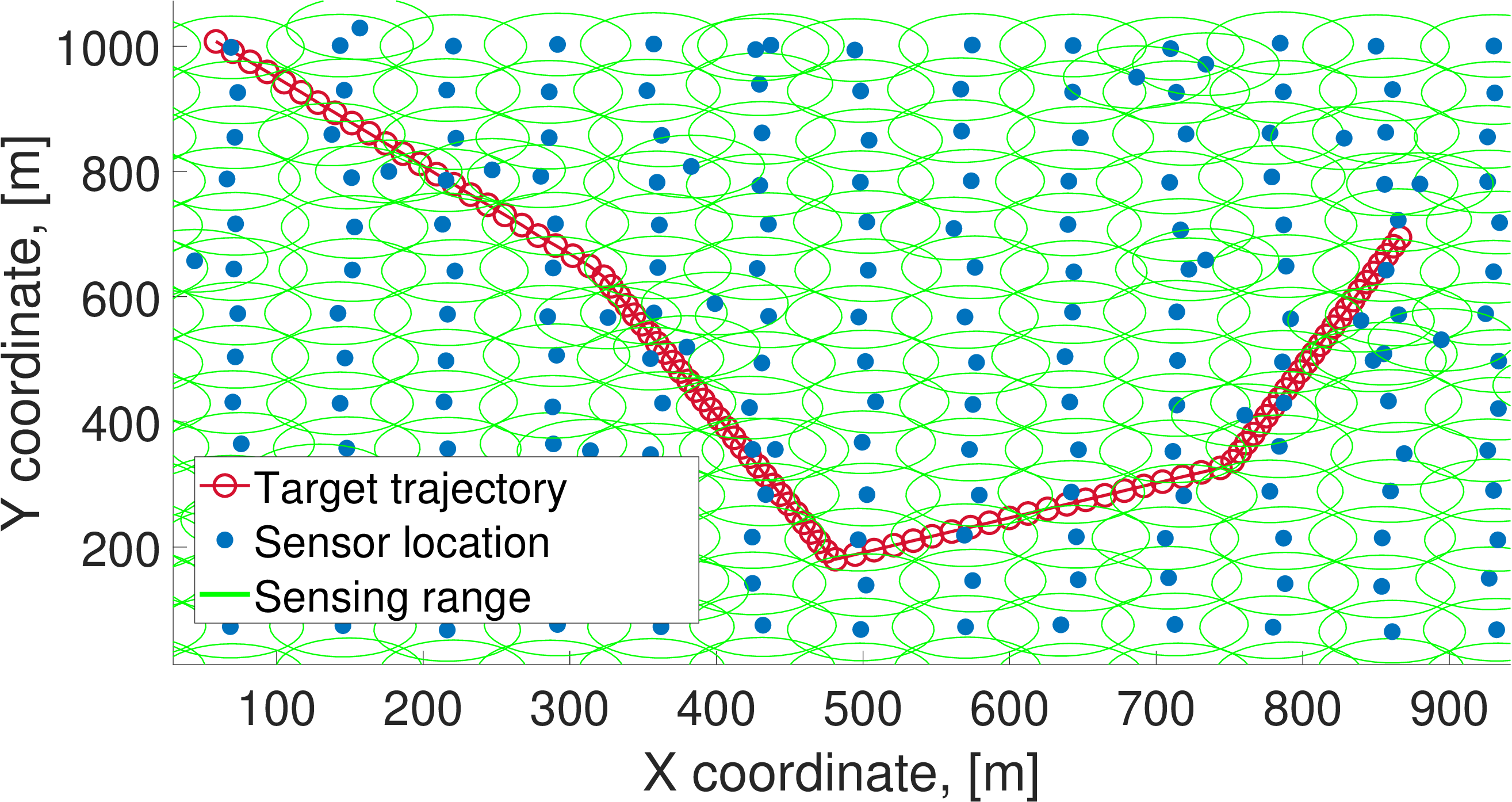}}
	\centerline{(a) Trajectory S1}\medskip
\end{minipage}
\hfill
\begin{minipage}[b]{0.49\linewidth}
	\centering
	\centerline{\includegraphics[width=4.35cm]{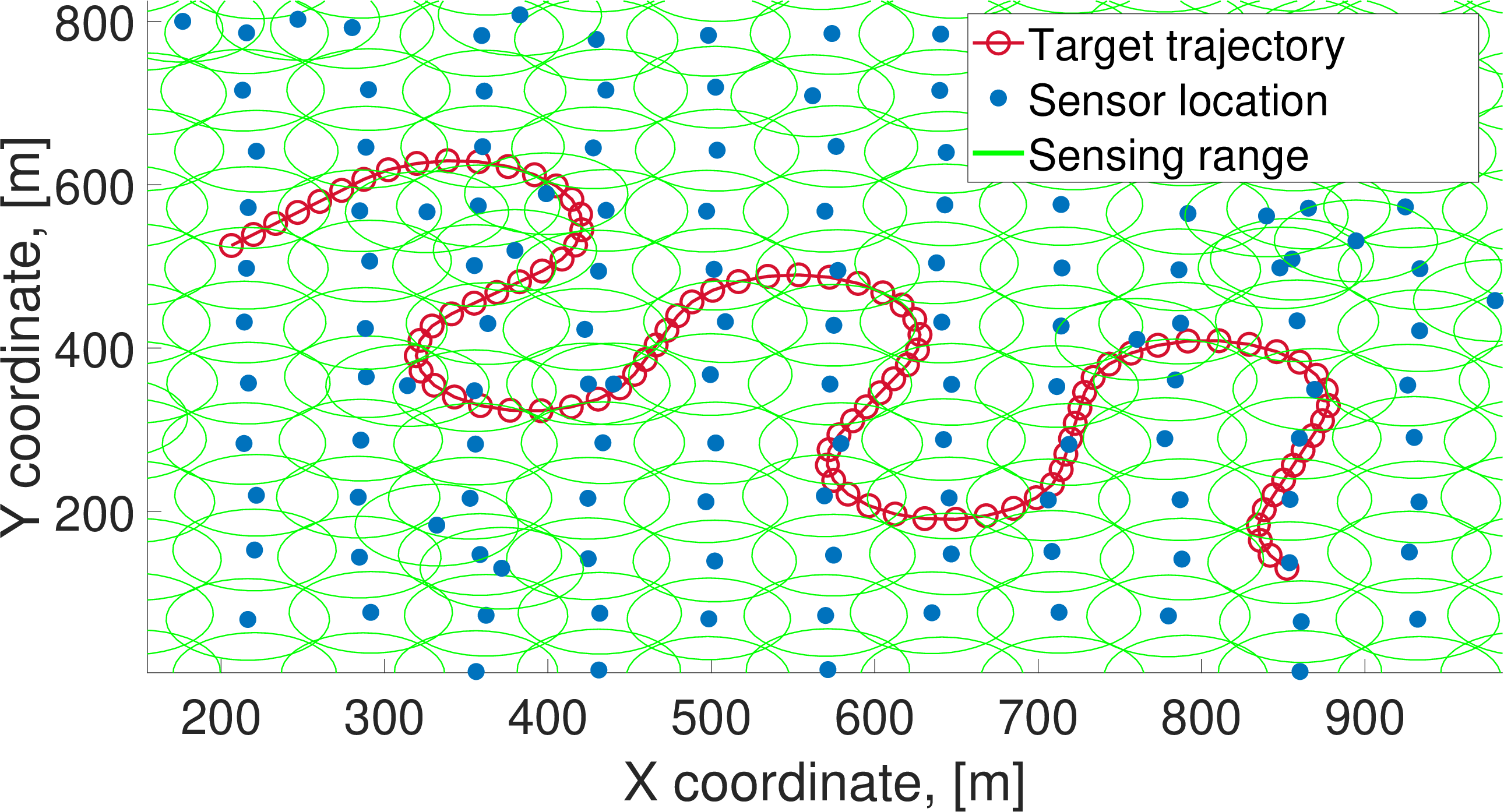}}
	\centerline{(b) Trajectory S2}\medskip
\end{minipage}
\begin{minipage}[b]{.49\linewidth}
  \centering
  \centerline{\includegraphics[width=4.35cm]{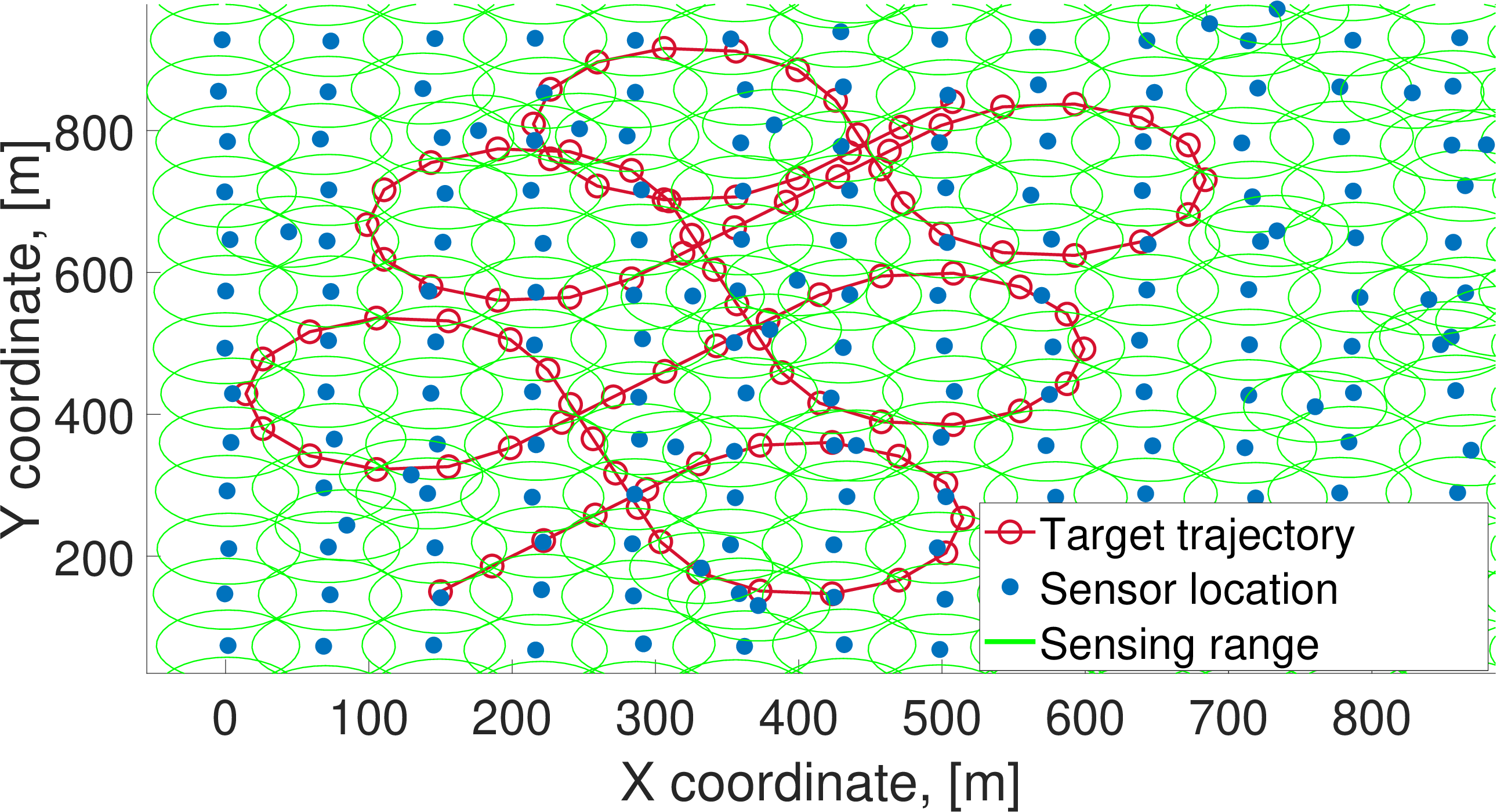}}
  \centerline{(c) Trajectory S3}\medskip
\end{minipage}
\hfill
\begin{minipage}[b]{0.49\linewidth}
  \centering
  \centerline{\includegraphics[width=4.35cm]{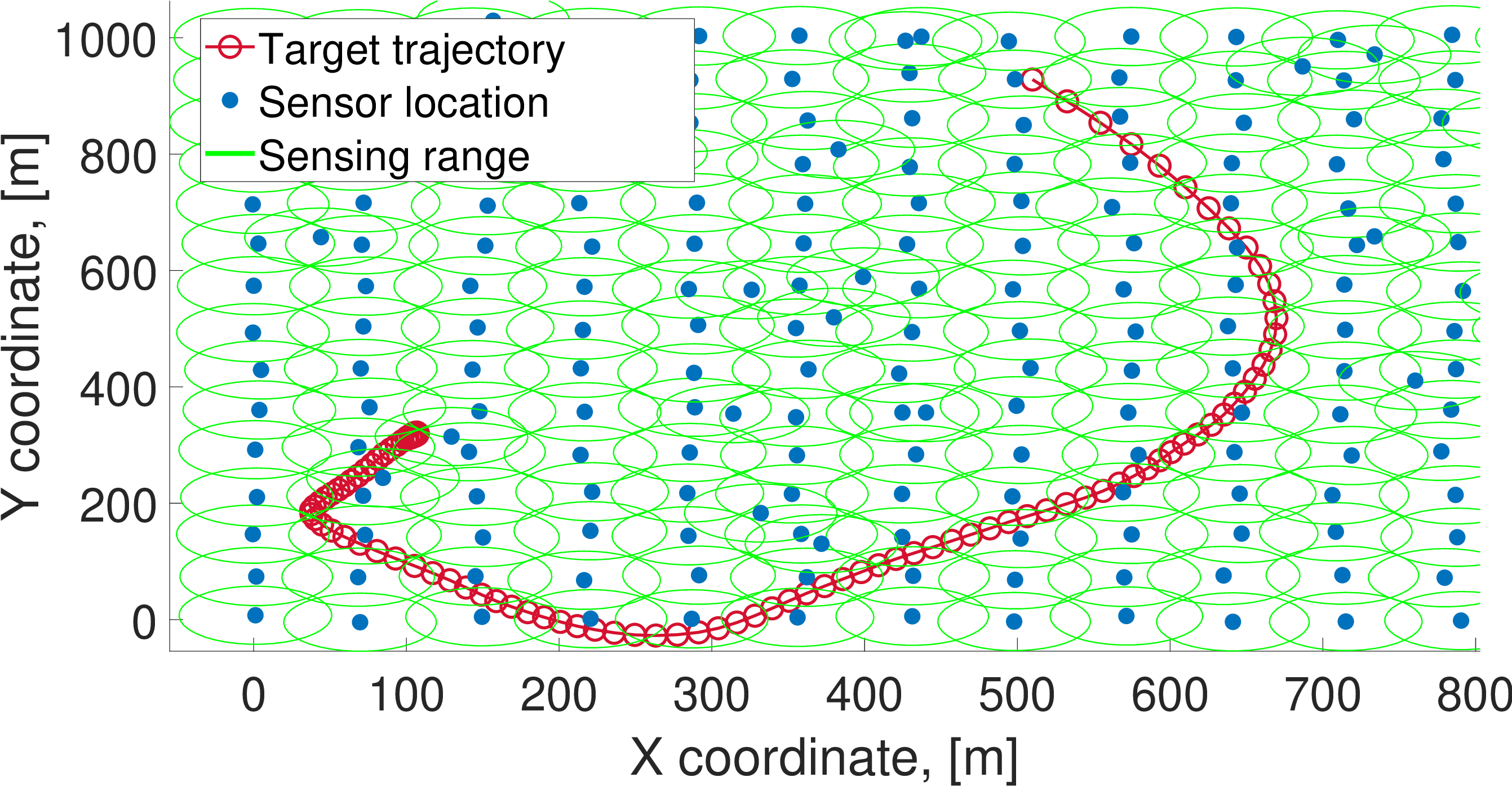}}
  \centerline{(d) Trajectory S4}\medskip
\end{minipage}
\caption{Target trajectories}
\label{fig3}
\end{figure}

To evaluate the proposed algorithm and the benchmarks, four challenging scenarios are built following different models. The trajectories and the sensors are depicted in Fig. \ref{fig3}.
\begin{itemize}
	\itemSa Similar to \cite{5203653}, the trajectory is generated based on the NCV model in the straight line, and the abrupt velocity change at each pre-defined turning point.
	\itemSb The target trajectory is generated by the gradual coordinated turns model ($20^{\circ}/s$ for 10 seconds) with the constant velocity model, which can go both left and right. This model can represent maneuvrable motions.
	\itemSc The target trajectory is generated by the sharp coordinated turns model which is more agile with higher turning rate ($30^{\circ}/s$ for 9 s).
	\itemSd The target trajectory is generated by the Singer acceleration model. The maximum possible acceleration is $50~m/s^2$, the probability of non-acceleration is $0.4$.
\end{itemize}

\subsection{Normalized Root Mean Square Errors}
In this section, the average normalized root mean square errors (NRMSEs) of the proposed DGPT and the standard GP-based tracker are evaluated. The NRMSE is defined as

\begin{align*}
    \text{NRMSE}=\sqrt{\frac{\sum\limits_i^N (\widehat{f}(\mathbf{x}_i)-f(\mathbf{x}_i))^2}{N}}\bigg/\left(f(\mathbf{x}_i)_\text{max}-f(\mathbf{x}_i)_\text{min}\right),
\end{align*}
where $f(\mathbf{x}_i)_\text{max}$ and $f(\mathbf{x}_i)_\text{min}$ represent the maximum and minimum value of the target states, respectively.

The tracking errors of the proposed DGPT approach with temporal input feature, which are collected from different trajectories are presented in Tables II-V. From the results, we can find that when considering the existence of clutter measurements, the clutter rate plays an important role in determining the prediction error. By contrast, the proposed DGPT approach achieves robust performance while changing the noise level of the target measurements. Moreover, using RBCM as the prediction aggregation method can achieve lower NRMSEs than GPoE in most scenarios, which justifies that adding the common prior into the aggregation process can improve the tracking accuracy. Particularly, the proposed DGPT approach performs competitively well and even outperforms the centralized approach in some scenarios. This is due to that in DGP, different weights can be assigned to the local predictions during the prediction aggregation process, so the final aggregated predictions are closer to the expert who makes more confident predictions. In the centralized method, all the data is aggregated before training without any difference.

The tracking errors of the proposed DGPT approach with both temporal and spatial input features are depicted in Tables VI-IX. We can find that considering spatial feature can help to improve the tracking accuracy as compared to TGP, especially in the more challenging scenarios where the speed of the target keeps changing or the target keeps maneuvering (Scenarios S2,s3, and S4). Particularly, the improvement is even more significant in the high clutter case since adding spatial feature can help to learn a more accurate likelihood function which can better assign weights for measurement preprocessing.
\begin{table}[!t]
	\centering\caption{Updated NRMSEs for S1: TGP}\label{Tab:1}
	\resizebox{0.49\textwidth}{!}{%
		\begin{tabular}{|c|c|c|c|c|c|c|c|}
			\hline
			& &\multicolumn{6}{c|}{Approach}    \\\cline{3-8} 
			Noise level&Clutter rate&\multicolumn{2}{c|}{Standard GP}  & \multicolumn{2}{c|}{DGPT-RBCM} & \multicolumn{2}{c|}{DGPT-GPoE}\\
			& & X & Y & X & Y &X & Y\\ \hline
			\multirow{2}{*}{1}& 1  & 0.77\% &  1.15\% &  0.93\% &  0.95\% & 1.51\% & 1.20\% \\ \cline{2-8}
			& 5  & 2.12\% &  1.78\% &  2.10\% &  2.14\% & 2.91\% & 2.72\% \\ \hline
			\multirow{2}{*}{2}& 1  & 0.79\% &  1.15\% &  0.94\% &  0.85\% & 1.60\% & 1.17\% \\ \cline{2-8}
			& 5  & 2.02\% &  1.94\% &  2.17\% &  2.57\% & 2.97\% & 3.11\% \\ \hline
			\multirow{2}{*}{3}& 1  & 0.84\% &  1.19\% &  0.98\% &  0.93\% & 1.70\% & 1.34\% \\ \cline{2-8}
			& 5  & 2.05\% &  1.91\% &  2.25\% &  3.01\% & 2.98\% & 3.59\% \\ \hline
		\end{tabular}%
	}
\end{table}

\begin{table}[!t]
	\centering\caption{Updated NRMSEs for S2: TGP}\label{Tab:2}
	\resizebox{0.49\textwidth}{!}{%
		\begin{tabular}{|c|c|c|c|c|c|c|c|}
			\hline
			& &\multicolumn{6}{c|}{Approach}    \\\cline{3-8} 
			Noise level&Clutter rate&\multicolumn{2}{c|}{Standard GP}  & \multicolumn{2}{c|}{DGPT-RBCM} & \multicolumn{2}{c|}{DGPT-GPoE}\\
			& & X & Y & X & Y &X & Y\\ \hline
			\multirow{2}{*}{1}& 1  & 1.22\% &  1.15\% &  1.50\% &  1.63\% & 2.04\% & 1.92\% \\ \cline{2-8}
			& 5  & 2.69\% &  2.75\% &  2.60\% &  3.00\% & 2.99\% & 3.27\% \\ \hline
			\multirow{2}{*}{2}& 1  & 1.25\% &  1.19\% &  1.43\% &  1.55\% & 1.85\% & 1.78\% \\ \cline{2-8}
			& 5  & 2.77\% &  2.85\% &  2.62\% &  3.33\% & 2.94\% & 3.65\% \\ \hline
			\multirow{2}{*}{3}& 1  & 1.32\% &  1.34\% &  1.55\% &  1.79\% & 1.87\% & 2.08\% \\ \cline{2-8}
			& 5  & 2.81\% &  2.90\% &  2.73\% &  3.50\% & 3.04\% & 3.87\% \\ \hline
		\end{tabular}%
	}
\end{table}

\begin{table}[!t]
	\centering\caption{Updated NRMSEs for S3: TGP}\label{Tab:3}
	\resizebox{0.49\textwidth}{!}{%
		\begin{tabular}{|c|c|c|c|c|c|c|c|}
			\hline
			& &\multicolumn{6}{c|}{Approach}    \\\cline{3-8} 
			Noise level&Clutter rate&\multicolumn{2}{c|}{Standard GP}  & \multicolumn{2}{c|}{DGPT-RBCM} & \multicolumn{2}{c|}{DGPT-GPoE}\\
			& & X & Y & X & Y &X & Y\\ \hline
			\multirow{2}{*}{1}& 1  & 1.82\% &  1.91\% &  1.81\% &  1.63\% & 1.82\% & 1.66\% \\ \cline{2-8}
			& 5  & 3.17\% &  3.40\% &  3.64\% &  3.23\% & 3.69\% & 3.29\% \\ \hline
			\multirow{2}{*}{2}& 1  & 1.81\% &  1.95\% &  1.84\% &  1.65\% & 1.85\% & 1.68\% \\ \cline{2-8}
			& 5  & 3.19\% &  3.42\% &  3.64\% &  3.21\% & 3.69\% & 3.27\% \\ \hline
			\multirow{2}{*}{3}& 1  & 1.86\% &  1.99\% &  1.88\% &  1.70\% & 1.90\% & 1.73\% \\ \cline{2-8}
			& 5  & 3.20\% &  3.43\% &  3.66\% &  3.23\% & 3.72\% & 3.29\% \\ \hline
		\end{tabular}%
	}
\end{table}

\begin{table}[!t]
	\centering\caption{Updated NRMSEs for S4: TGP}\label{Tab:4}
	\resizebox{0.49\textwidth}{!}{%
		\begin{tabular}{|c|c|c|c|c|c|c|c|}
			\hline
			& &\multicolumn{6}{c|}{Approach}    \\\cline{3-8} 
			Noise level&Clutter rate&\multicolumn{2}{c|}{Standard GP}  & \multicolumn{2}{c|}{DGPT-RBCM} & \multicolumn{2}{c|}{DGPT-GPoE}\\
			& & X & Y & X & Y &X & Y\\ \hline
			\multirow{2}{*}{1}& 1  & 0.98\% &  0.70\% &  0.87\% &  0.56\% & 1.13\% & 0.73\% \\ \cline{2-8}
			& 5  & 1.64\% &  1.59\% &  2.77\% &  1.90\% & 3.88\% & 2.53\% \\ \hline
			\multirow{2}{*}{2}& 1  & 0.99\% &  0.71\% &  1.04\% &  0.61\% & 1.37\% & 0.80\% \\ \cline{2-8}
			& 5  & 1.75\% &  1.48\% &  2.83\% &  1.92\% & 3.95\% & 2.58\% \\ \hline
			\multirow{2}{*}{3}& 1  & 1.05\% &  0.73\% &  1.36\% &  0.79\% & 1.97\% & 1.10\% \\ \cline{2-8}
			& 5  & 1.87\% &  1.48\% &  3.02\% &  2.02\% & 4.21\% & 2.67\% \\ \hline
		\end{tabular}%
	}
\end{table}

\begin{table}[!t]
	\centering\caption{Updated NRMSEs for S1: STGP}\label{Tab:5}
	\resizebox{0.49\textwidth}{!}{%
		\begin{tabular}{|c|c|c|c|c|c|c|c|}
			\hline
			& &\multicolumn{6}{c|}{Approach}    \\\cline{3-8} 
			Noise level&Clutter rate&\multicolumn{2}{c|}{Standard GP}  & \multicolumn{2}{c|}{DGPT-RBCM} & \multicolumn{2}{c|}{DGPT-GPoE}\\
			& & X & Y & X & Y &X & Y\\ \hline
			\multirow{2}{*}{1}& 1  & 0.73\% &  0.80\% &  0.81\% &  0.82\% & 1.27\% & 1.02\% \\ \cline{2-8}
			& 5  & 1.93\% &  1.52\% &  2.64\% &  1.98\% & 3.31\% & 2.18\% \\ \hline
			\multirow{2}{*}{2}& 1  & 0.73\% &  0.81\% &  0.84\% &  0.84\% & 1.35\% & 1.03\% \\ \cline{2-8}
			& 5  & 1.92\% &  1.59\% &  2.68\% &  2.07\% & 3.34\% & 2.25\% \\ \hline
			\multirow{2}{*}{3}& 1  & 0.80\% &  0.88\% &  0.93\% &  0.91\% & 1.59\% & 1.09\% \\ \cline{2-8}
			& 5  & 1.95\% &  1.59\% &  2.64\% &  2.04\% & 3.24\% & 2.28\% \\ \hline
		\end{tabular}%
	}
\end{table}

\begin{table}[!t]
	\centering\caption{Updated NRMSEs for S2: STGP}\label{Tab:6}
	\resizebox{0.49\textwidth}{!}{%
		\begin{tabular}{|c|c|c|c|c|c|c|c|}
			\hline
			& &\multicolumn{6}{c|}{Approach}    \\\cline{3-8} 
			Noise level&Clutter rate&\multicolumn{2}{c|}{Standard GP}  & \multicolumn{2}{c|}{DGPT-RBCM} & \multicolumn{2}{c|}{DGPT-GPoE}\\
			& & X & Y & X & Y &X & Y\\ \hline
			\multirow{2}{*}{1}& 1  & 1.07\% &  1.38\% &  1.29\% &  1.67\% & 2.48\% & 2.15\% \\ \cline{2-8}
			& 5  & 2.02\% &  2.21\% &  2.10\% &  2.44\% & 3.31\% & 2.86\% \\ \hline
			\multirow{2}{*}{2}& 1  & 1.10\% &  1.40\% &  1.30\% &  1.72\% & 2.54\% & 2.13\% \\ \cline{2-8}
			& 5  & 2.01\% &  2.19\% &  2.14\% &  2.47\% & 3.41\% & 2.82\% \\ \hline
			\multirow{2}{*}{3}& 1  & 1.15\% &  1.47\% &  1.35\% &  1.71\% & 2.62\% & 2.09\% \\ \cline{2-8}
			& 5  & 2.03\% &  2.18\% &  2.18\% &  2.46\% & 3.26\% & 2.82\% \\ \hline
		\end{tabular}%
	}
\end{table}

\begin{table}[!t]
	\centering\caption{Updated NRMSEs for S3: STGP}\label{Tab:7}
	\resizebox{0.49\textwidth}{!}{%
		\begin{tabular}{|c|c|c|c|c|c|c|c|}
			\hline
			& &\multicolumn{6}{c|}{Approach}    \\\cline{3-8} 
			Noise level&Clutter rate&\multicolumn{2}{c|}{Standard GP}  & \multicolumn{2}{c|}{DGPT-RBCM} & \multicolumn{2}{c|}{DGPT-GPoE}\\
			& & X & Y & X & Y &X & Y\\ \hline
			\multirow{2}{*}{1}& 1  & 3.25\% &  2.93\% &  1.42\% &  1.24\% & 1.47\% & 1.21\% \\ \cline{2-8}
			& 5  & 2.91\% &  4.09\% &  2.41\% &  1.85\% & 2.49\% & 1.86\% \\ \hline
			\multirow{2}{*}{2}& 1  & 3.43\% &  2.93\% &  1.66\% &  1.19\% & 1.66\% & 1.23\% \\ \cline{2-8}
			& 5  & 2.83\% &  4.41\% &  2.26\% &  1.87\% & 2.56\% & 1.94\% \\ \hline
			\multirow{2}{*}{3}& 1  & 3.64\% &  2.95\% &  1.44\% &  1.21\% & 1.47\% & 1.28\% \\ \cline{2-8}
			& 5  & 2.78\% &  3.98\% &  2.35\% &  1.86\% & 2.41\% & 1.94\% \\ \hline
		\end{tabular}%
	}
\end{table}

\begin{table}[!t]
	\centering\caption{Updated NRMSEs for S4: STGP}\label{Tab:8}
	\resizebox{0.49\textwidth}{!}{%
		\begin{tabular}{|c|c|c|c|c|c|c|c|}
			\hline
			& &\multicolumn{6}{c|}{Approach}    \\\cline{3-8} 
			Noise level&Clutter rate&\multicolumn{2}{c|}{Standard GP}  & \multicolumn{2}{c|}{DGPT-RBCM} & \multicolumn{2}{c|}{DGPT-GPoE}\\
			& & X & Y & X & Y &X & Y\\ \hline
			\multirow{2}{*}{1}& 1  & 1.49\% &  0.96\% &  1.46\% &  0.94\% & 1.93\% & 1.11\% \\ \cline{2-8}
			& 5  & 2.77\% &  1.32\% &  2.75\% &  1.53\% & 2.99\% & 1.63\% \\ \hline
			\multirow{2}{*}{2}& 1  & 1.48\% &  0.97\% &  1.47\% &  0.90\% & 1.93\% & 1.13\% \\ \cline{2-8}
			& 5  & 2.81\% &  1.30\% &  2.58\% &  1.49\% & 3.00\% & 1.61\% \\ \hline
			\multirow{2}{*}{3}& 1  & 1.61\% &  0.99\% &  1.51\% &  0.97\% & 2.01\% & 1.12\% \\ \cline{2-8}
			& 5  & 2.81\% &  1.31\% &  2.64\% &  1.49\% & 2.94\% & 1.55\% \\ \hline
		\end{tabular}%
	}
\end{table}

\subsection{Uncertainty Quantification}
\begin{figure}[t]
    \centering
	\includegraphics[width=3.7in]{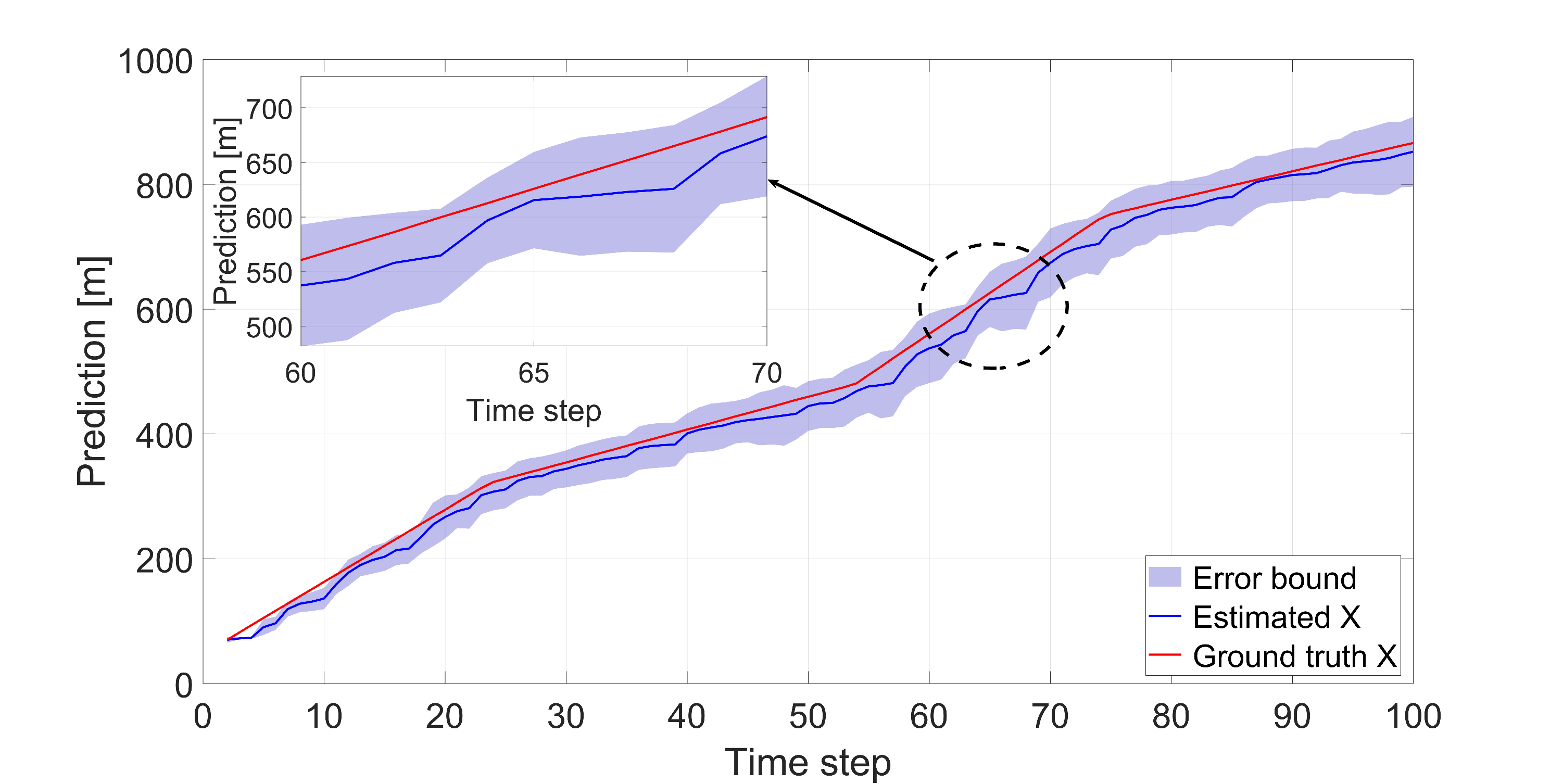}
	\caption{UCB for DGPT-RBCM of S1 in X coordinate}\label{error1}
\end{figure}
\begin{figure}[t]
    \centering
	\includegraphics[width=3.7in]{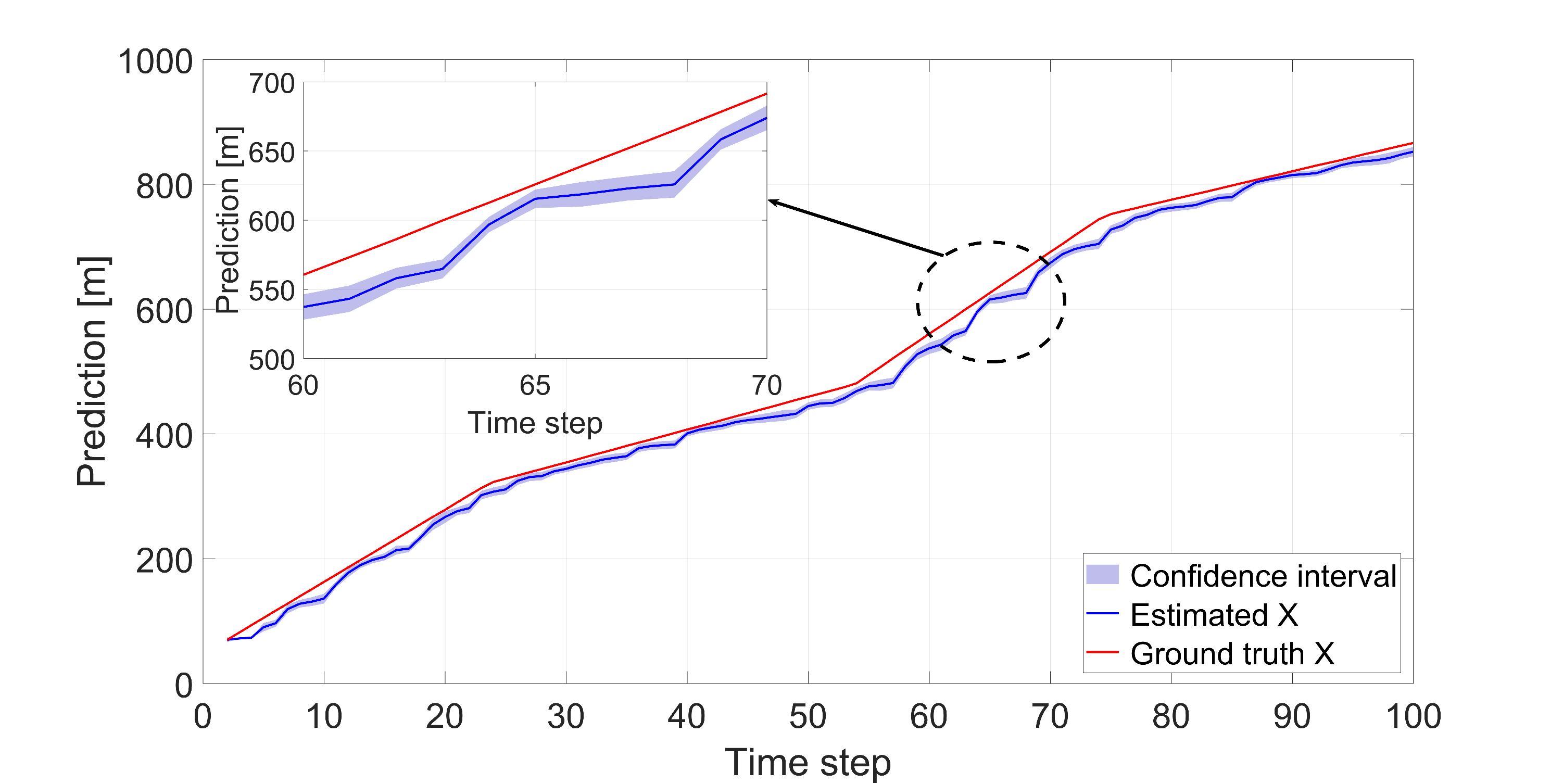}
	\caption{Confidence interval for DGPT-RBCM of S1 in X coordinate}\label{error3}
\end{figure}
\begin{figure}[t]
    \centering
	\includegraphics[width=3.7in]{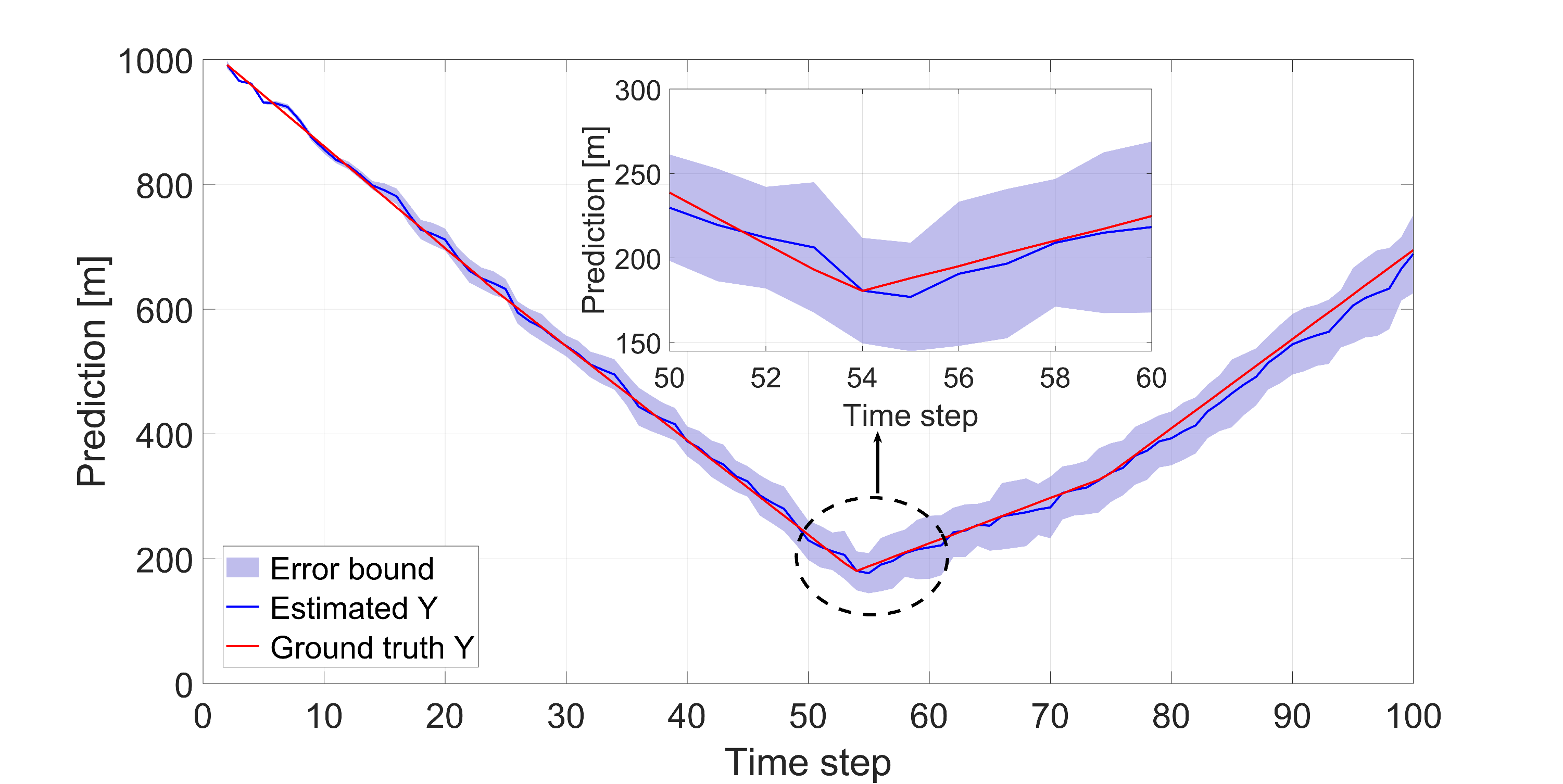}
	\caption{UCB for DGPT-RBCM of S1 in Y coordinate}\label{error2}
\end{figure}
\begin{figure}[t]
    \centering
	\includegraphics[width=3.7in]{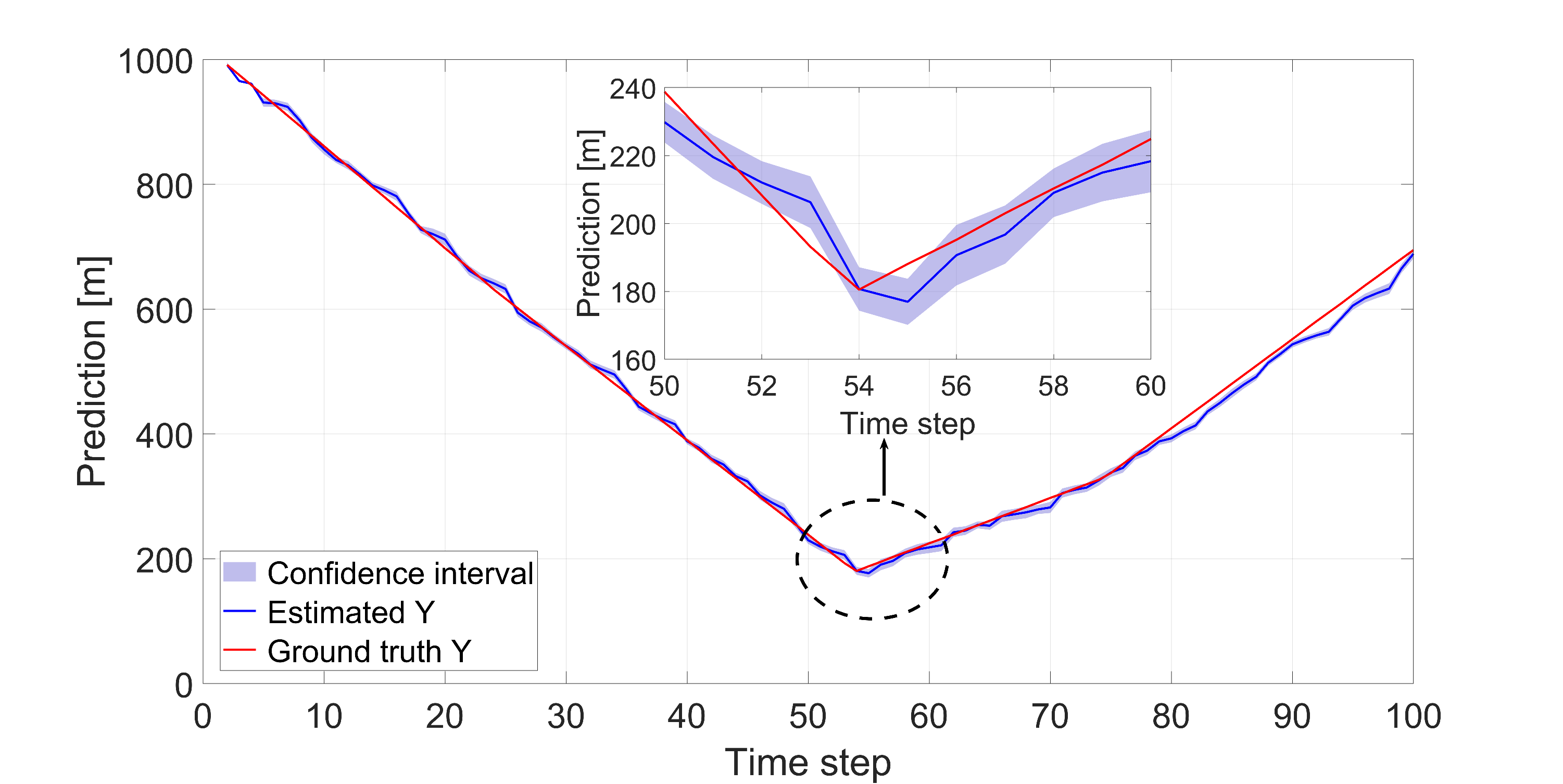}
	\caption{Confidence interval for DGPT-RBCM of S1 in Y coordinate}\label{error4}
\end{figure}
To visualise the derived UCB of the proposed DGPT, we choose the probability that the UCB holds as 99.7\% which corresponds to the 3$\sigma$ confidence interval of the Gaussian distribution. Based on theoretical analysis from Theorem 2, the UCBs of predictions of DGPT-RBCM in both X and Y coordinates are presented in Figures \ref{error1} and \ref{error2}. The confidence intervals of the predictive distribution of the DGPT-RBCM in both X and Y coordinates based on \eqref{RBCM2} and \eqref{RBCM3} are presented in Figures \ref{error3} and \ref{error4}. The results in Figures \ref{error1} and \ref{error2} demonstrate that the proposed UCB can reveal the information of where the true location of the target is since the UCB can cover the true target location in most time steps with a high probability. However, in Figures \ref{error3} and \ref{error4}, although the confidence intervals can quantify the uncertainties of the predictions themselves, the intervals fail to cover the true location of the target in most of the time steps. The derived UCB better characterizes the presence of the target in the error bound with $88\%$ and $42\%$ higher probability in $X$ and $Y$ coordinates, as compared to the confidence of DGP. The comparisons between the derived UCB and the confidence interval highlight the value and informativeness of the UCB, which can help to further refine the measurements for training the DGP model by excluding some measurements out of the bound and thus having the potential to further enhance the measurement preprocessing method presented in Section \ref{sec:4.3}. 
\subsection{Hybrid Bayesian Filtering}
\begin{figure*}[!t]
\begin{minipage}[b]{.55\linewidth}
	\centering
	\centerline{\includegraphics[width=7cm]{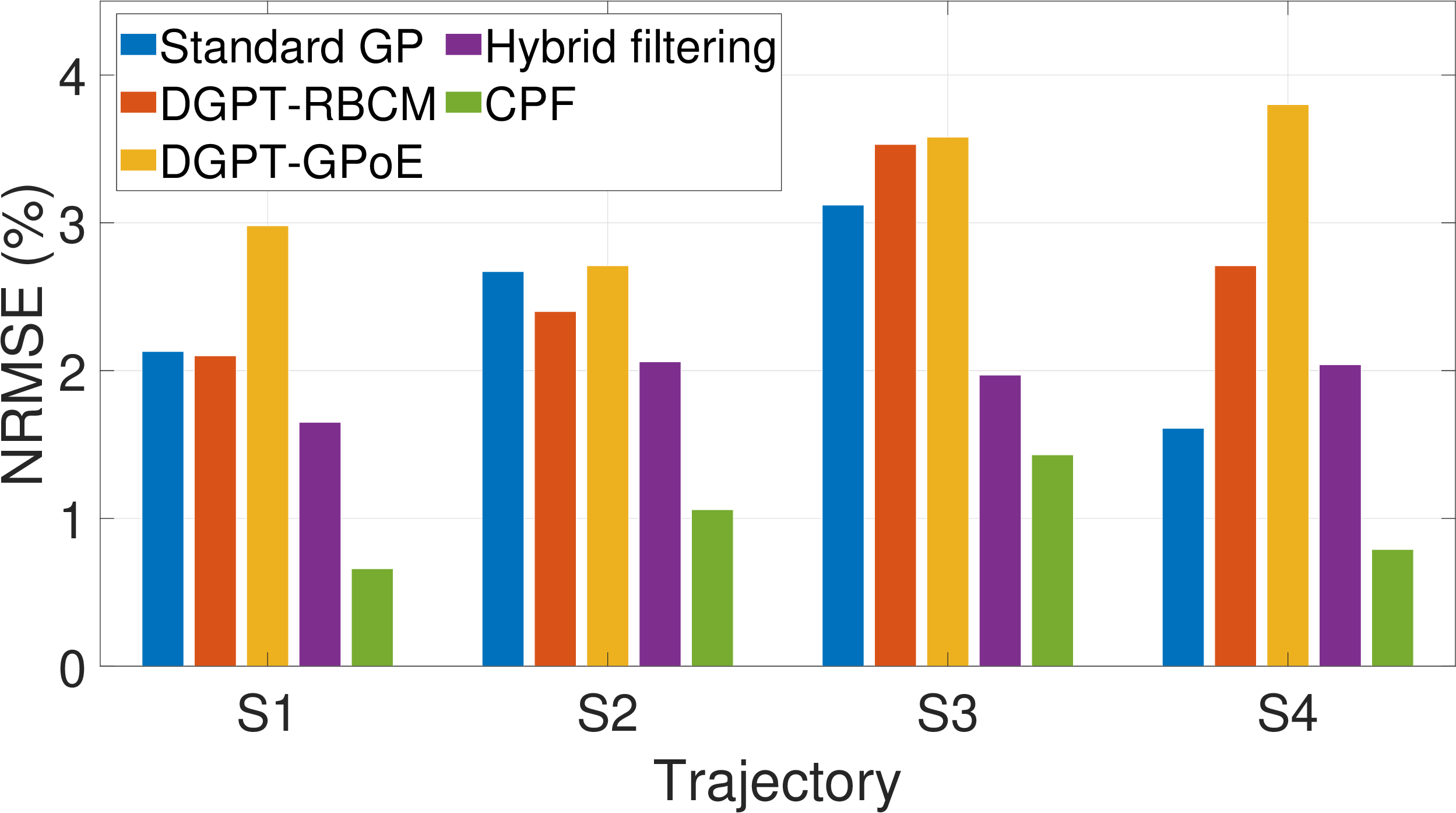}}
	\centerline{(a) NRMSE: X coordinate}\medskip
\end{minipage}
\hfill
\begin{minipage}[b]{-.55\linewidth}
	\centering
	\centerline{\includegraphics[width=7cm]{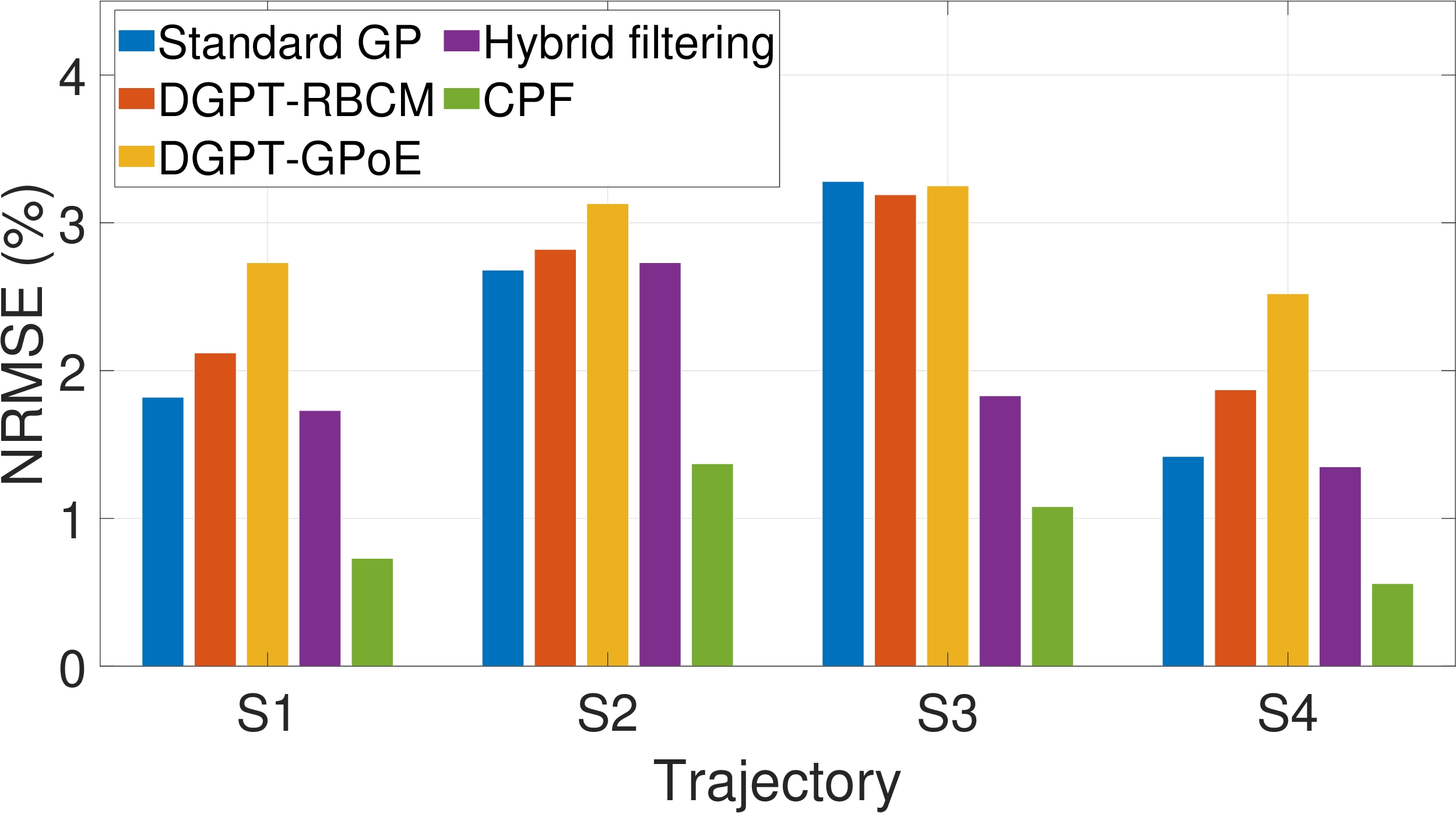}}
	\centerline{(b) NRMSE: Y coordinate}\medskip
\end{minipage}
\begin{minipage}[b]{.55\linewidth}
  \centering
  \centerline{\includegraphics[width=7cm]{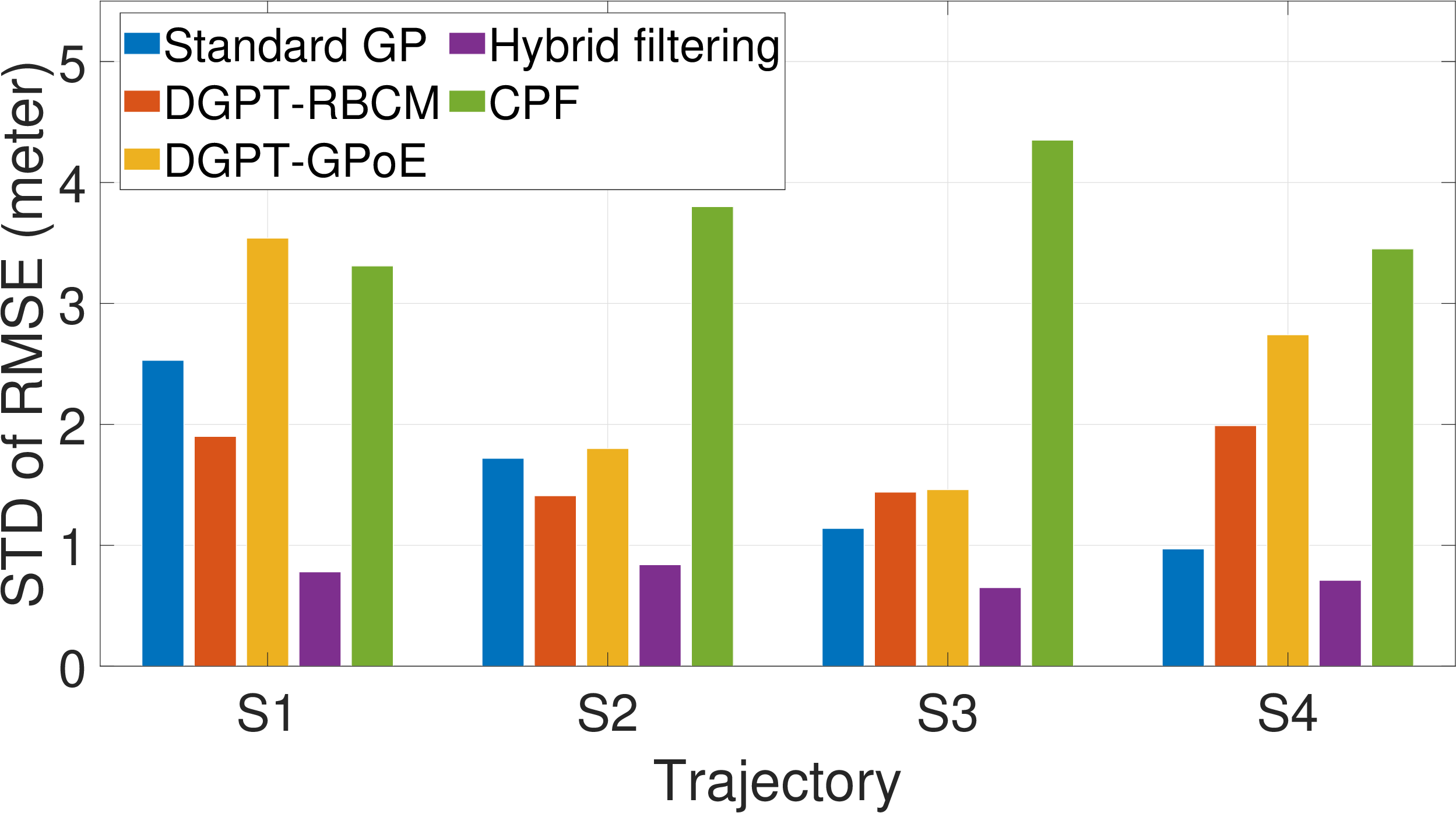}}
  \centerline{(c) STD of the RMSE: X coordinate}\medskip
\end{minipage}
\hfill
\begin{minipage}[b]{-.55\linewidth}
  \centering
  \centerline{\includegraphics[width=7cm]{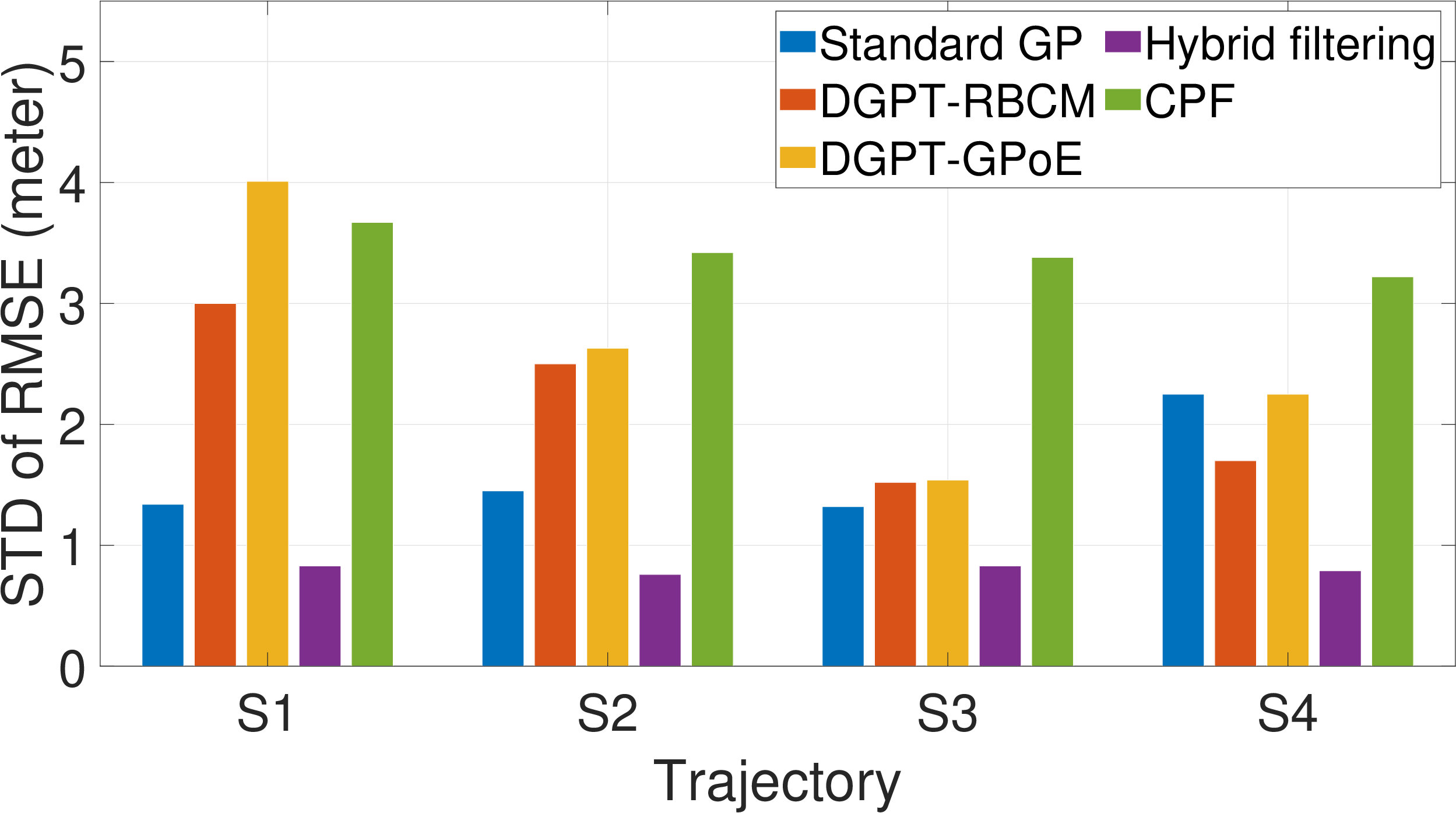}}
  \centerline{(d) STD of the RMSE: Y coordinate}\medskip
\end{minipage}
\caption{Tracking error and the STD of errors over 100 MC runs, $\sigma^2_z=1$, clutter rate=5}
\label{fig_Hybrid_Bayesia_Filtering}
\end{figure*}
In this section, the performance of the proposed hybrid Bayesian filtering approach for target tracking is evaluated as compared to both the Standard GP-based tracker and the DGPT approach. {\color{black}We have also compared with a model-based aapproach by using a convolutional particle filter (CPF)~\cite{6641185,Doucet2001,musso2001improving,6335120,defreitasetal:2016:autonomouscrowdstracking} which is one of the best versions of the particle filters. The CPF is implemented by processing the data in a centralized manner. 

The CPF has two types of models - one constant velocity model and 4 coordinated turn models. A grid of values for the angular turn rate $\omega$ is chosen as
\begin{align}
	\omega=\left\{ 0,0.55,0.49,-0.55,-0.49\right\} s^{-1}.
\end{align}
These values cover possible uniform motions and maneuver within the minimum and maximum values of the turn rate. The CPF multiple model filter is tested over all 4 considered scenarios, by keeping the same transition probabilities matrix, initial mode probabilities vector and grid for the angular turn values. This means that we we give the same conditions for the CPF, for all testing scenarios. 

Inspired by ideas from Gilholm and Salmond \cite{Gilholm1,gilholm2005spatial}, we adopt the powerful Poisson likelihood model
in the CPF and in our proposed hybrid Bayesian filtering approach in order to deal with measurement origin uncertainties and solve the data association task.

The Poisson likelihood model for dealing with measurement origin uncertainties leads to one of the most efficient data association approaches which avoids combinatorial complexities which are typical for multiple hypothesis target tracking model-based approaches. 

The dynamic model with a known turn rate has the following system matrix:
\[F(\omega)=
\begin{bmatrix}
	1 & \sin \omega T_s/\omega & 0 &  -(1-\cos\omega T_s)/\omega \\
	0 & \cos \omega T_s& 0 &  -\sin \omega T_s \\
	0 & (1-\cos\omega T_s)/\omega & 1 &  \sin \omega T_s/\omega \\
	0 & \sin \omega T_s & 0 &  \cos \omega T_s
\end{bmatrix},
\]
where $T_s$ is the sampling interval and we set $T_s=1$. The value of  $\omega=0$ corresponds to the nonmaneuvering (constant velocity) model. The values $\pm \omega$ correspond to the left and right turn, respectively. The transition mode probabilities are assumed to follow a Markov chain, with the initial mode probability as $\mu_0=0.6, \mu_1=0.1, \mu_2=0.1, \mu_3=0.1, \mu_4=0.1$. The mode transition matrix is defined as
\[
\begin{bmatrix}
	P_{1,1} & P_{1,2} & P_{1,3} &  P_{1,4} &  P_{1,5} \\
	P_{2,1} & P_{2,2} & P_{2,3} &  P_{2,4} &  P_{2,5} \\
	P_{3,1} & P_{3,2} & P_{3,3} &  P_{3,4} &  P_{3,5} \\
	P_{4,1} & P_{4,2} & P_{4,3} &  P_{4,4} &  P_{4,5} \\
	P_{5,1} & P_{5,2} & P_{5,3} &  P_{5,4} &  P_{5,5} \\
\end{bmatrix}\]

\[=
\begin{bmatrix}
	0.84 & 0.05 & 0.03 &  0.05 &  0.03 \\
	0.05 & 0.88 & 0.07 &  0 &  0 \\
	0.03 & 0.07 & 0.9 &  0 &  0 \\
	0.05 & 0 & 0 &  0.88 &  0.07 \\
	0.03 & 0 & 0 &  0.07 &  0.9 \\
\end{bmatrix}.
\]


We choose the challenging case where the clutter rate is set as $\lambda_{\text{c}}=5$ (high clutter case), and the measurement noise is set to be $\sigma^2_z=1$. The normalized root mean squared errors (NRMSEs) of the $X$ and $Y$ coordinates over 100 MC runs are given in Figures A (a) and (b). The hybrid Bayesian filtering approach outperforms both the centralized and distributed GP-based tracking approaches by achieving the lowest NRMSEs of both coordinates (except the NRMSE in S4, $X$ coordinate). In particular, we can see that a fine-tuned multiple-model CPF achieves the lowest NRMSE values in both $X$ and $Y$ coordinates.

We also present the standard deviation (STD) of averaged RMSEs over 100 MC runs. Each of the RMSEs is the tracking error calculated from a single independent MC run. The results are presented in Figures A (c) and (d). The proposed hybrid Bayesian filtering approach achieves the lowest STD of state estimations in both coordinates as compared to other GP-based approaches. It is also fairly stable over multiple scenarios. The results demonstrate that by involving the Poisson likelihood model, both the accuracy and the robustness of the proposed DGPT can be further improved. The STDs of RMSEs from the CPF are relatively higher in some cases, which shows the proposed algorithm achieves higher robustness than the model-based tracking approach. The high STDs of the RMSEs for the CPF also mean that the estimates from the CPF are somehow far from the true target trajectories in some areas or some runs. }



\section{Conclusion}
In this paper, a novel DGP-based model-free learning and tracking approach is proposed to solve distributed point tracking problems in WSNs with clutter measurements. The developed approach belongs to the distributed edge learning and overcomes the limitations of standard GP-based tracking methods from a different angle via distributed GP. Theoretical derivations are presented for the UCB of the tracking error for two important tasks: 1) when data have no clutter, 2) when the sensor data contain clutter which means there is measurement origin uncertainty. The UCBs characterize the trustworthiness of the proposed approach. The estimates are acceptable when the derived UCB is within certain pre-specified limits. It characterizes the presence of the target in the error bound with $88\%$ and $42\%$ higher probability in $X$ and $Y$ coordinates, respectively, than the confidence interval-based method. By introducing the Poisson measurement likelihood, a hybrid Bayesian filtering approach is proposed to merge the distributed machine learning and model-based methods to further improve the tracking performance and robustness. Numerical experiments demonstrate that the proposed approaches perform competitively well and can deal with varying motion models, noise levels, and clutter rates. {\color{black}Future work will focus on sensor management challenges in the developed distributed tracking system. The challenges include efficiently utilizing edge computing resources to accelerate DGP training and prediction. Another direction is to consider different types of measurements such as received signal strength and non-Gaussian measurement noises in the distributed tracking system. The theoretical derivation of a PCRLB is an important topic to focus on, and the validation of the theoretical results over real case studies will also be considered in the future.}

\vspace{-3mm}
\section*{Appendix A}\label{sec:appendix_A}
According to \eqref{RBCM1}, \eqref{RBCM2}, Lemma 1 and the analysis as in Theorem 1, the deviation of the true function value from the aggregated predictive mean by RBCM can be written as
\begin{align}
    &\lvert f(\mathbf{x}_*)-\mu^\text{RBCM}_* \rvert,\nonumber\\=&\lvert f(\mathbf{x}_*)-\frac{\sum^M_{i=1}\beta_i\sigma^{-2}_i(\mathbf{x}_*)\mu_i(\mathbf{x}_*)}{\sum^M_{i=1}\beta_i\sigma^{-2}_i(\mathbf{x}_*)+(1-M)\sigma^{-2}_{**}}\rvert,\label{bound3_1}\\
    \approx & \Bigg\lvert f(\mathbf{x}_*)-\frac{\sum^M_{i=1}\beta_i\sigma^{-2}_i(\mathbf{x}_*)\mu_i(\mathbf{x}_*)}{\sum^M_{i=1}\beta_i\sigma^{-2}_i(\mathbf{x}_*)}\Bigg\rvert,\label{bound3_2}\\
    \leq & \frac{\sum^M_{i=1}\beta_i\gamma_i^{1/2}\sigma^{-1}_i(\mathbf{x}_*)}{\sum^M_{i=1}\beta_i\sigma^{-2}_i(\mathbf{x}_*)}.
\end{align}
Equations (\ref{bound3_1})-(\ref{bound3_2}) are derived when $\sigma_{**}^2$ (the variance of the prior distribution) has a big value which corresponds to lack of prior information or high uncertainty.


\section*{Appendix B}\label{sec:appendix_B}
Based on Lemma \ref{lemma2}, similar to Theorem 1, define $B_i$ as the event that for any $t\in T$, the cumulative prediction of the target state from local expert $i$ and the true target state differs larger than a quantity, which can be written as
\begin{align}
    B_i=\left\{\bigcup\nolimits_{t=1}^T\lvert f(\mathbf{x}_t)-\mu^\text{GPoE}(\mathbf{x}_t) \rvert >\gamma_{t,i}^{1/2}\sigma_{i}(\mathbf{x}_t)\right\},
\end{align}
where $\mu_{i}(\mathbf{x}_t)$ and $\sigma_{i}(\mathbf{x}_t)$ represents the predictive mean and the STD of the function made by local expert $i$ at time $t$, respectively.

Define the union of events $\left\{B_1,B_2,\cdots,B_M\right\}$ as $B$. By applying the union bounds over $M$ events, the probability of $B$ can be upper bounded as
\begin{align}
    \text{Pr}(B)&\defeq\text{Pr}\left\{ \bigcup\nolimits_{i=1}^{M} B_i \right\}\leq \sum\nolimits_{i=1}^M \delta_i.
\end{align}

Therefore, we have
\begin{align}
    \text{Pr}(\Bar{B})&\defeq\text{Pr}\left\{ \bigcap\nolimits_{i=1}^{M} \Bar{B_i} \right\}\nonumber,\\&=\text{Pr}\left\{ \bigcap\nolimits_{i=1}^{M} \bigcap\nolimits_{t=1}^{T}\lvert f(\mathbf{x}_t)-\mu_{i}(\mathbf{x}_t) \rvert \leq\gamma_{t,i}^{1/2}\sigma_{i}(\mathbf{x}_t) \right\}\nonumber,\\&\geq 1-\sum\nolimits_{i=1}^M \delta_i.
\end{align}

Therefore, with probability $1-\sum_{i=1}^M \delta_i$, we have
\begin{align}
    &\sum_{t=1}^T\lvert f(\mathbf{x}_t)-\mu^\text{GPoE}_t \rvert \nonumber \\=& \sum_{t=1}^T\lvert f(\mathbf{x}_t)-\frac{\sum^M_{i=1}\beta_{t,i}\sigma^{-2}_{i}(\mathbf{x}_t)\mu_{i}(\mathbf{x}_t)}{\sum^M_{i=1}\beta_{t,i}\sigma^{-2}_{i}(\mathbf{x}_t)}\rvert\nonumber,\\
    \leq&\sum_{t=1}^T\frac{\sum^M_{i=1}\beta_{t,i}\sigma^{-2}_{i}(\mathbf{x}_t)\lvert f(\mathbf{x}_t)-\mu_{i}(\mathbf{x}_t)\rvert}{\sum^M_{i=1}\beta_{t,i}\sigma^{-2}_{i}(\mathbf{x}_t)}\nonumber,\\
    \leq&\sum_{t=1}^T\frac{\sum^M_{i=1}\beta_{t,i}\gamma_{t,i}^{1/2}\sigma^{-1}_{i}(\mathbf{x}_t)}{\sum^M_{i=1}\beta_{t,i}\sigma^{-2}_{i}(\mathbf{x}_t)}.
\end{align}

\section*{Appendix C}\label{sec:appendix_C}
The derivation of the expressions for the first two moments of the posterior state distribution is presented as follows:
\newpage
\begin{strip}
  \begin{align}
   p(\tilde{\mathbf{X}}_t|\mathbf{Z}_{t-C:t})~\propto~& p(\mathbf{z}_t|\tilde{\mathbf{X}}_t)p(\tilde{\mathbf{X}}_t|\mathbf{Z}_{t-C:t-1}) ,\\
    =~& p(z_{t,1}|\tilde{\mathbf{X}}_t)(z_{t,2}|\tilde{\mathbf{X}}_t)\cdots(z_{t,n_t}|\tilde{\mathbf{X}}_t)p(\tilde{\mathbf{X}}_t|\mathbf{Z}_{t-C:t-1})\\
    =~&\prod_{j=1}^{n_t}\left\{\frac{1}{\sqrt{2\pi \hat{\sigma}_{t}^2}}\exp{\frac{-(z_{t,j}-\hat{\mu}_t)^2}{2\hat{\sigma}_{t}^2}}\right\}\cdot\frac{1}{\sqrt{2\pi \sigma_{t-1}^2}}\exp{\frac{-(\tilde{x}_t-\mu_{t-1})^2}{2\sigma_{t-1}^2}},\\
    \propto~&\exp{\left\{ \frac{-\sum_{j=1}^{n_t}(z_{t,j}^2-2z_{t,j}\hat{\mu}_t+\hat{\mu}_t^2)}{2\hat{\sigma}_{t}^2} - \frac{\tilde{x}_t^2-2\tilde{x}_t\mu_{t-1}+\mu_{t-1}^2}{2\sigma_{t-1}^2} \right\}},\\
    \propto~&\exp{\left\{ \frac{-\sigma_{t-1}^2\sum_{j=1}^{n_t}\left(z_{t,j}^2-2z_{t,j}(\frac{\lambda_{\text{c}}}{A_\text{sen}}+\lambda_{\text{T}} \tilde{x}_t)+(\frac{\lambda_{\text{c}}}{A_\text{sen}}+\lambda_{\text{T}} \tilde{x}_t)^2\right)-\hat{\sigma}_{t}^2(\tilde{x}_t^2-2\tilde{x}_t\mu_{t-1}+\mu_{t-1}^2)}{2\hat{\sigma}_{t}^2\sigma_{t-1}^2} \right\}},\\
    \propto~&\exp{\left\{ \frac{-\tilde{x}_t^2(\hat{\sigma}_{t}^2+n_t \sigma_{t-1}^2\lambda_{\text{T}}^2)+2\tilde{x}_t\left( \mu_{t-1}\hat{\sigma}_{t}^2+\sigma_{t-1}^2\sum_{j=1}^{n_t}(z_j\lambda_{\text{T}}-\lambda_{\text{T}}\lambda_{\text{c}}/A_\text{sen})\right)}{2\hat{\sigma}_{t}^2\sigma_{t-1}^2} \right\}},\\
    \propto~&\exp{\left\{  \frac{-\tilde{x}_t^2+2\tilde{x}_t\left(\frac{\mu_{t-1}\hat{\sigma}_{t}^2+\sigma_{t-1}^2\sum_{j=1}^{n_t}(z_j\lambda_{\text{T}}-\lambda_{\text{T}}\lambda_{\text{c}}/A_\text{sen})}{\hat{\sigma}_{t}^2+n_t \sigma_{t-1}^2\lambda_{\text{T}}^2} \right)}{2\frac{\hat{\sigma}_{t}^2\sigma_{t-1}^2}{\hat{\sigma}_{t}^2+n_t \sigma_{t-1}^2\lambda_{\text{T}}^2}}  \right\}},\\
    \propto~&\exp{\left\{ \frac{-\left(\tilde{x}_t-\frac{\mu_{t-1}\hat{\sigma}_{t}^2+\sigma_{t-1}^2\sum_{j=1}^{n_t}(z_j\lambda_{\text{T}}-\lambda_{\text{T}}\lambda_{\text{c}}/A_\text{sen})}{\hat{\sigma}_{t}^2+n_t \sigma_{t-1}^2\lambda_{\text{T}}^2} \right)^2}{2\frac{\hat{\sigma}_{t}^2\sigma_{t-1}^2}{\hat{\sigma}_{t}^2+n_t \sigma_{t-1}^2\lambda_{\text{T}}^2}} \right\}}.
  \end{align}
\end{strip}
\bibliographystyle{IEEEtran}
\bibliography{Reference}

\end{document}